\documentclass[lettersize,journal]{IEEEtran}
\usepackage{amsmath,amsfonts}
\usepackage{algorithmic}
\usepackage{algorithm}
\usepackage{array}
\usepackage[caption=false,font=normalsize,labelfont=rm,textfont=rm]{subfig}
\usepackage{textcomp}
\usepackage{stfloats}
\usepackage{url}
\usepackage{verbatim}
\usepackage{graphicx}
\usepackage{cite}
\usepackage{amsmath}
\usepackage{bbding}
\usepackage{latexsym}
\usepackage{svg}
\usepackage{pifont}
\usepackage{amssymb}
\usepackage{booktabs}
\usepackage{multirow}
\usepackage{multicol}
\usepackage{caption}
\usepackage{cite}
\usepackage{subfig}
\usepackage{multirow}
\usepackage{booktabs}
\usepackage{bm}
\usepackage{tablefootnote}
\usepackage[table]{xcolor}
\usepackage[
  colorlinks=true,       
  linkcolor=red,         
  citecolor=green,        
  filecolor=magenta,    
  urlcolor=cyan,         
  bookmarks=true,        
  pdfborder={0 0 0}      
]{hyperref}
\newcommand{\red}[1]{\textcolor{black}{#1}}
\newcommand{\blue}[1]{\textcolor{black}{#1}}

\usepackage[capitalize]{cleveref}
\crefname{section}{Sec.}{Secs.}
\Crefname{section}{Section}{Sections}
\Crefname{table}{Table}{Tables}
\crefname{table}{Tab.}{Tabs.}
\newcommand{\ie}{i.e.,\ }

\newcommand{\etal}{\textit{et al.}}

\begin{document}

\title{Dual Learning with Dynamic Knowledge Distillation and Soft Alignment for Partially Relevant Video Retrieval
}

\author{Jianfeng~Dong,
        Lei~Huang,
        Daizong~Liu,
        Xianke~Chen,
        Xun~Yang,
        Changting~Lin,\\
        Xun~Wang,~\IEEEmembership{Member,~IEEE,}
        and~Meng~Wang,~\IEEEmembership{Fellow,~IEEE,}%
        
\thanks{J. Dong, L. Huang, X. Chen, and X. Wang are with the College of Computer and Information Engineering, Zhejiang Gongshang University, Hangzhou 310035, China (e-mail: dongjf24@gmail.com, huangxixiyiqi@gmail.com, cxk\_zjgsu@163.com, wx@zjgsu.edu.cn).}%

\thanks{D. Liu is with the Wangxuan Institute of Computer Technology, Peking University, Beijing 100871, China (e-mail: dzliu@stu.pku.edu.cn).}%

\thanks{X. Yang is with the School of Information Science and Technology, University of Science and Technology of China, Hefei 230026, China (e-mail: xyang21@ustc.edu.cn).}%

\thanks{C. Lin is with the Binjiang Institute of Zhejiang University, Hangzhou, China, and with GenTel.io, Hangzhou, China (e-mail: linchangting@gmail.com).}%

\thanks{M. Wang is with the School of Computer Science and Information Engineering, Hefei University of Technology, Hefei 230009, China (e-mail: wangmeng@hfut.edu.cn).}%

\thanks{Corresponding author: Xianke Chen.}
}



\maketitle

\begin{abstract}
Almost all previous text-to-video retrieval works ideally assume that videos are pre-trimmed with short durations containing solely text-related content. However, in practice, videos are typically untrimmed in long durations with much more complicated background content. Therefore, in this paper, we focus on the more practical yet challenging task of Partially Relevant Video Retrieval (PRVR), which aims to retrieve partially relevant untrimmed videos with the given query. 
To tackle this task, we propose a novel framework that distills generalization knowledge from a powerful large-scale vision-language pre-trained model and transfers it to a lightweight, task-specific PRVR network. Specifically, we introduce a Dual Learning framework with Dynamic Knowledge Distillation (DL-DKD++), where a large teacher model provides supervision to a compact dual-branch student network. The student model comprises two branches: an inheritance branch that absorbs transferable knowledge from the teacher, and an exploration branch that learns task-specific information from the PRVR dataset to address domain gaps.
To further enhance learning, we incorporate a dynamic soft-target construction mechanism. By replacing rigid hard-target supervision with adaptive soft targets that evolve during training, our method enables the model to better capture the fine-grained, partial relevance between videos and queries.
Experiment results demonstrate that our proposed model achieves state-of-the-art performance on TVR, ActivityNet, and Charades-STA datasets for PRVR. 
The code is available at \href{https://github.com/HuiGuanLab/DL-DKD}{https://github.com/HuiGuanLab/DL-DKD}.
\end{abstract}

\begin{IEEEkeywords}
Text-to-video Retrieval, Partially Relevant Video Retrieval, Knowledge Distillation, Soft Alignment.
\end{IEEEkeywords}

\section{Introduction}
\label{sec:intro}

\begin{figure}[tb!]
    \subfloat[]{
    \includegraphics[width=0.57\columnwidth]{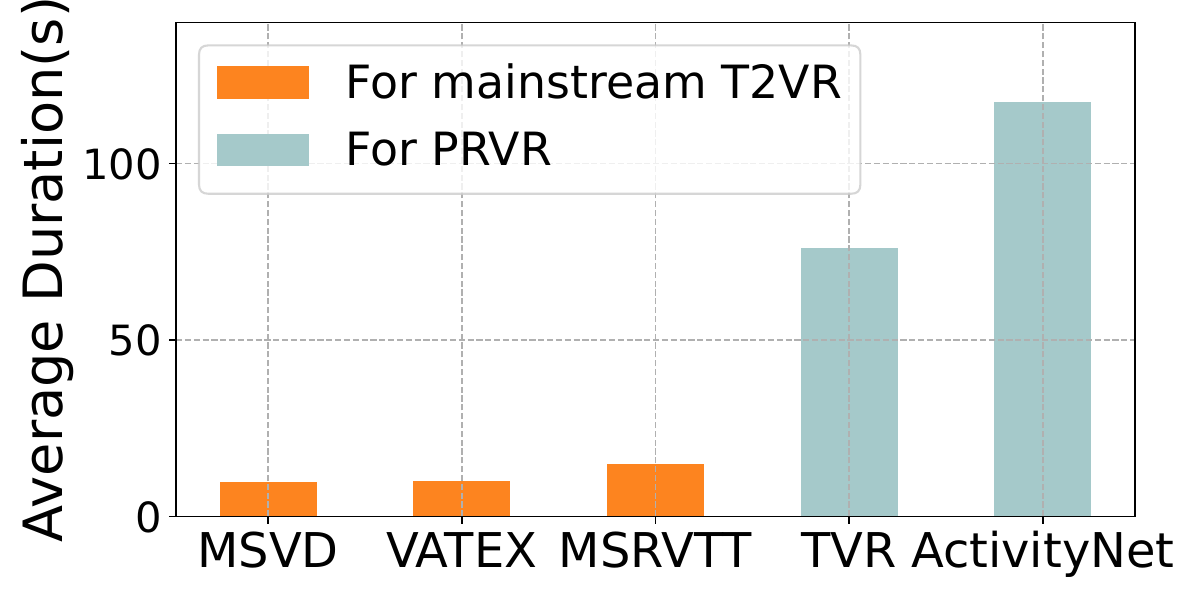}
    }
    \subfloat[]{
    \includegraphics[width=0.37\columnwidth]{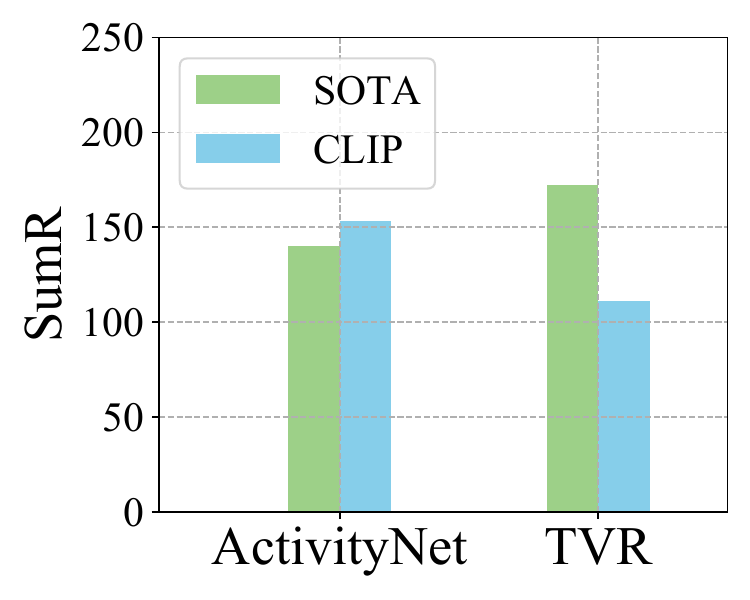}
    }
    \newline
    \subfloat[]{
    \includegraphics[width=0.97\columnwidth]{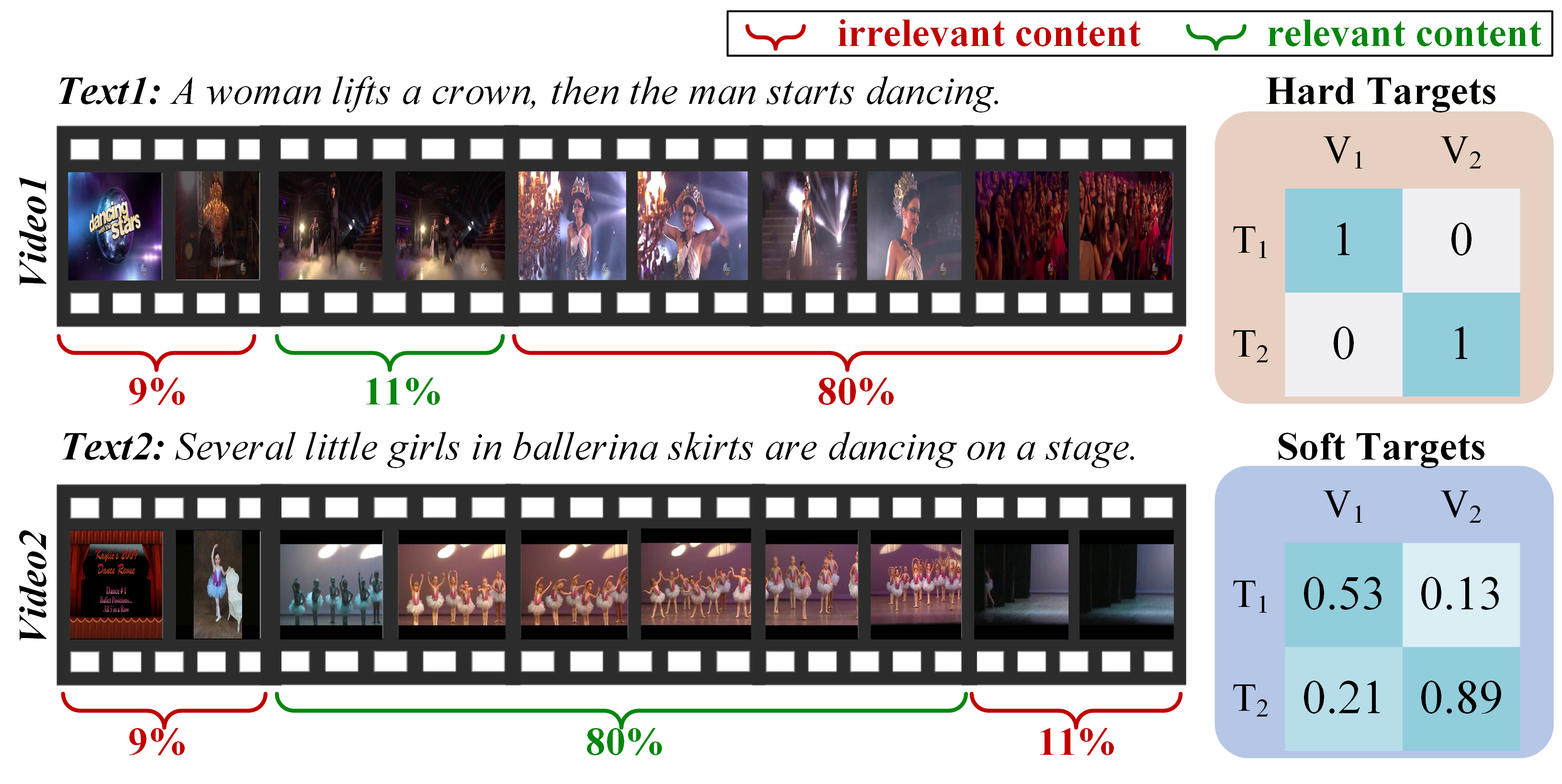}
    }

    \caption{
    (a) Compared to mainstream T2VR dataset, PRVR datasets involve much longer untrimmed videos;
    (b) Performance comparisons between the state-of-the-art (SOTA)~\cite{dong2022prvr} and the vanilla CLIP-based\cite{max2022guide}  methods, where CLIP shows huge performance divergences for different PRVR datasets;
    (c) Partial relevance inherently varies in degree, and soft targets better capture this characteristic than hard binary targets.
    }
    
    \vspace{-3mm}
    \label{fig:dynamic_distillation}
\end{figure}

\noindent
\IEEEPARstart{W}{ith} the explosion of online videos, searching for videos of interest has been an indispensable activity in people’s daily lives.
In this context, text-to-video retrieval (T2VR), 
retrieving videos \textit{w.r.t.} a textual query from a large number of unlabeled videos, attracts growing attention recently~\cite{peng2017overview,xu2020joint,li2022adaptive,wang2022wang,gao2023vectorized}.  A fundamental assumption underlying most mainstream T2VR approaches is that videos are pre-trimmed into short segments that are entirely relevant to the query~\cite{yu2018joint,li2020sea,song2021spatial}. 
While this assumption simplifies the modeling process, it does not reflect real-world scenarios, where videos are often untrimmed, lengthy, and interspersed with irrelevant or noisy content.
In such scenarios, queries may only correspond to a small portion of the video, and pre-trimmed clips may omit important contextual information. This mismatch between the assumptions of conventional T2VR models and practical video retrieval demands highlights a critical need for more flexible retrieval paradigms.

To fill this gap, a new text-to-video retrieval subtask, \textit{i.e.}, Partially Relevant Video Retrieval (PRVR), has been proposed recently in~\cite{dong2022prvr}. Different from conventional T2VR, PRVR aims to retrieve the partially relevant untrimmed videos that contain at least one internal moment relevant to the given query, while potentially also including irrelevant content.
This allows the videos used for PRVR to be much longer than those for T2VR (see Fig. \ref{fig:dynamic_distillation}(a)), being more practical in daily usage. 
Therefore, in this work, we target on the PRVR task, considering it is more consistent with practical video retrieval scenarios.

Recently, large-scale pre-trained vision-language models, such as Contrastive Language-Image Pre-training (CLIP) \cite{radford2021learning}, have shown great promise across various cross-modal tasks, including T2VR. 
Since CLIP is trained on image-text pairs without explicitly modeling temporal dependencies, most CLIP-based video retrieval methods rely on developing temporal aggregation layers to handle the inherent sequential structure of video data~\cite{liu2022transfer,max2022guide,luo2022clip4clip,fang2021clip2video}.
While effective for short pre-trimmed videos, these methods struggle in PRVR due to the presence of mixed relevant and irrelevant segments within long untrimmed videos.
Simply applying mainstream T2VR methods to PRVR by aggregating CLIP features across all frames can result in substantial performance degradation. As shown in Fig.~\ref{fig:dynamic_distillation}(b), a vanilla CLIP performs well on ActivityNet but depressing on TVR.
This underscores a critical challenge: how to effectively transfer CLIP’s knowledge to PRVR models remains an open research problem.

To address these issues, we propose a novel Dual Learning framework with Dynamic Knowledge Distillation (DL-DKD++) that adaptively purifies and transfers the knowledge of CLIP into the PRVR with task-specific supervision.
Our framework adopts a teacher-student architecture, with CLIP as the large teacher and a lightweight dual-branch student model as the learner.
The two student branches serve complementary roles:
(1) an inheritance branch focuses on directly absorbing useful knowledge from CLIP for a specific domain, and
(2) an exploration branch is used to learn task-specific patterns from the training data, addressing potential domain gaps.
Additionally, motivated by the way humans first learn from teachers and then gradually engage in self-directed learning as their own understanding develops, we devise a dynamic knowledge distillation strategy. Initially, the inheritance branch takes precedence, while the exploration branch becomes more prominent throughout the training process. In this manner, our DL-DKD++ is able to take advantage of both the powerful generalization ability of CLIP and the benefits of task-specific model convergence on the PRVR data.

Additionally, we observe that previous PRVR methods \cite{dong2022prvr,wang2024gmmformer} adopt a binary supervision strategy with hard targets, where a video is labeled as positive only if it contains content exactly relevant to the textual query. This binary assumption oversimplifies the nature of partial relevance in untrimmed videos, neglecting the duration of relevant content in videos. 
In practice, different videos may contain the relevant content to varying degrees.
For instance, as illustrated in Fig.~\ref{fig:dynamic_distillation}(c), two videos depict the action of \textit{dancing}, while they differ in relevance based on the duration and prominence of the action. 
Treating them equally fails to distinguish the strength of their semantic alignment with the query.
To alleviate this, we propose a dynamic soft-target supervision mechanism that better captures the partial relevance of PRVR. 
In the early stage of training, we keep hard targets that provide clear and stable guidance to help the model establish initial discriminative capabilities. However, as the model matures, the binary hard targets become limiting, as they cannot capture the varying degrees of partial relevance crucial for PRVR. Therefore, we gradually incorporate soft targets of model-predicted similarity scores into the supervision, allowing the model to evolve toward flexible, fine-grained relevance modeling.

To sum up, the contributions of this work are as follows: 
\begin{itemize}
    \item We propose a novel knowledge distillation framework that contains a dual-branch student network to acquire appropriate knowledge selectively for partially relevant video retrieval. \red{Meanwhile, our framework supports single-teacher and multiple-teacher distillation.}
    \item We explore how to take the advantage of the powerful generalization-ability of the large model and the benefits of the task-specific model simultaneously while alleviating their limitations, and propose a dynamical knowledge distillation strategy.
    \item 
    We propose a dynamic soft-target construction module that generates flexible soft supervision for both positive and negative samples, addressing the limitations of binary hard target strategies in PRVR.
    \item Extensive experiments on three datasets demonstrate the effectiveness of our proposed framework, and it achieves new state-of-the-art performance on the challenging PRVR task.
\end{itemize}

A preliminary version of this work was published at ICCV 2023~\cite{dong2023dual}. This journal extension brings significant improvements over the conference version in three main aspects.
(1) Conceptually, we emphasize that partial relevance in PRVR is inherently continuous rather than binary, motivating the use of soft targets to better model partial similarity and capture varying degrees of relevance.
(2) Technically, we introduce a dynamic soft-target construction module that assigns graded relevance levels to samples during training, enabling the model to more effectively capture the varying degrees of partial relevance inherent in PRVR.
(3) Experimentally, we strengthen the evaluation by adding an additional benchmark dataset (Charades-STA) and incorporating several new competitive baselines for more comprehensive comparisons. We further conduct detailed ablation studies to isolate the contribution of the dynamic soft-target module. To provide deeper insight, we also present qualitative visualizations and data distribution analyses that illustrate the behavior and benefits of our proposed approach.

\section{Related Work} \label{sec:rel-work}

\subsection{Text-to-video Retrieval}

Given a textual query, the task of T2VR~\cite{wei2021universal,yang2020tree,liu2021context,liu2020jointly} aims to retrieve relevant videos with the query from a set of pre-trimmed video clips.
The dominant methods typically project videos and queries into a common space for measuring the cross-modal similarity~\cite{liu2019use,feng2021exploiting,gabeur2020multi,wu2021hanet}. They usually learn the cross-modal similarity using a large amount of video-text pairs, based on the initial video features extracted by pre-trained vision models and the text features obtained by pre-trained language models\cite{9741388}.
Additionally, we observe an increasing use of large-scale pre-training vision-language models, such as CLIP~\cite{radford2021learning}, for text-to-video retrieval~\cite{liu2022transfer,max2022guide,luo2022clip4clip,hu2022lightweight,fang2021clip2video}.
For instance, Hu \etal \cite{hu2022lightweight} utilize CLIP to extract both video and text features as extra features.
Other works adapt CLIP for text-to-video retrieval by introducing the similarity calculation module between the representation of text and video frames~\cite{luo2022clip4clip}, frame-wise attentions~\cite{max2022guide}, and a temporal difference block for capturing motions between frames~\cite{fang2021clip2video}, \cite{Liu_2025_CVPR}.

In practice, videos are generally untrimmed containing much background content \cite{qu2020fine,zheng2023progressive,xiao2020visual,ji2023binary}. However, in the traditional T2VR, videos are typically pre-trimmed with short duration and are supposed to be fully relevant to the query~\cite{yu2018joint,song2021spatial,gao2021fast}, which leads to a huge gap between the literature and the real world. To overcome this limitation, a new text-to-video retrieval subtask, \ie, PRVR, has been proposed~\cite{dong2022prvr}. By contrast,  videos in PRVR are typically untrimmed, and it aims to retrieve partially relevant videos with the query. An untrimmed video is considered to be partially relevant to a given textual query if it contains a moment relevant to the query. Although it is more consistent with real applications, PRVR had been neglected for a long time. 

\subsection{Partially Relevant Video Retrieval.}
Unlike conventional T2VR, the PRVR task assumes that each text query may relevant only to a specific moment within an untrimmed video. This formulation aligns well with real-world applications, highlighting its practical importance. As both the location and duration of the relevant moment are unknown, recent works formulate the task as a multiple instance learning problem, which typically represent videos as sequences of clips and define partial relevance as the maximum similarity between the query and any of the clips~\cite{dong2022prvr,wang2024gmmformer,nishimuravision}.
As a pioneering work, Dong \etal \cite{dong2022prvr} were the first to formally investigate the PRVR task, a dual-branch architecture that estimates partial relevance at both the clip and frame levels in a coarse-to-fine manner was proposed. But they use an explicit sliding window to generate clips, which leads to redundancy and high storage cost despite its effectiveness.
Building upon this foundation but aiming for greater efficiency, Wang et al.~\cite{wang2024gmmformer} generate clips through a Gaussian mixture model based Transformer, avoiding the need for explicit sliding windows. To further improve discriminability, they also introduce a query diverse loss to mitigate embedding sparsity by enhancing separation between queries linked to the same video.
While also focusing on enhancing temporal modeling precision, Chen et al.~\cite{chen2023joint} introduce an alternative dual-branch framework. Their approach leverages proposal generation and Gaussian-weighted pooling to capture temporal relevance more precisely. Taking a significantly different approach to modeling, Nishimura et al.~\cite{nishimuravision} convert entire videos into spatial \textit{super images} and reformulate PRVR as an image-text retrieval problem.

Despite recent progress, existing PRVR methods still face two key limitations: They solely learn from the target PRVR dataset without extra prior knowledge, which limits their ability to handle fine-grained or ambiguous queries~\cite{gao2023vectorized,10605104,10153964}; They use binary hard-target supervision, which fails to capture the varying degrees of partial relevance across video-query pairs.
To overcome these limitations, we propose to leverage knowledge from large pre-trained vision-language models to provide rich semantic priors, and introduce a dynamic soft-target supervision mechanism to enable flexible and fine-grained partial relevance modeling.

\subsection{Knowledge Distillation}

Knowledge distillation is the process of transferring knowledge from a large model (teacher) to a smaller one (student) \cite{hinton2015distilling,hu2024preface,yang2019distillhash,huang2018deep}, which has been widely employed in various tasks, such as image classification~\cite{mirzadeh2020improved}, and object detection~\cite{chen2017learning}. 
The distillation methods are generally categorized into two types: logit distillation \cite{hinton2015distilling} and feature distillation \cite{chen2021distilling}.
Logit-based knowledge distillation \cite{hinton2015distilling} methods aligns the teacher and student models by optimizing their distributions of the similarities between video-text samples. Feature-based knowledge distillation methods \cite{chen2021distilling}, \cite{10509807} provide additional guidance to the student model by optimizing its intermediate activation maps to resemble those of the teacher model.
Recently, a number of works demonstrate that knowledge distillation is also beneficial to learning text-to-video retrieval models~\cite{croitoru2021teachtext,miech2021thinking,fang2022learning,Mi_2022_CVPR,10446485}. 
For instance, Croitoru \etal \cite{croitoru2021teachtext} distill multiple strong text encoders into one text encoder. Miech \etal \cite{miech2021thinking} aims to learn an efficient text-to-visual retrieval model via distilling from a cross-attention model of high performance to a dual encoder model. These works usually have a prerequisite that teacher models have pretty good performance.
In this work, we relax this prerequisite, allowing teacher models of mediocrity to be used for knowledge distillation.

\begin{figure*}[htbp]
	\includegraphics[width=1\linewidth]{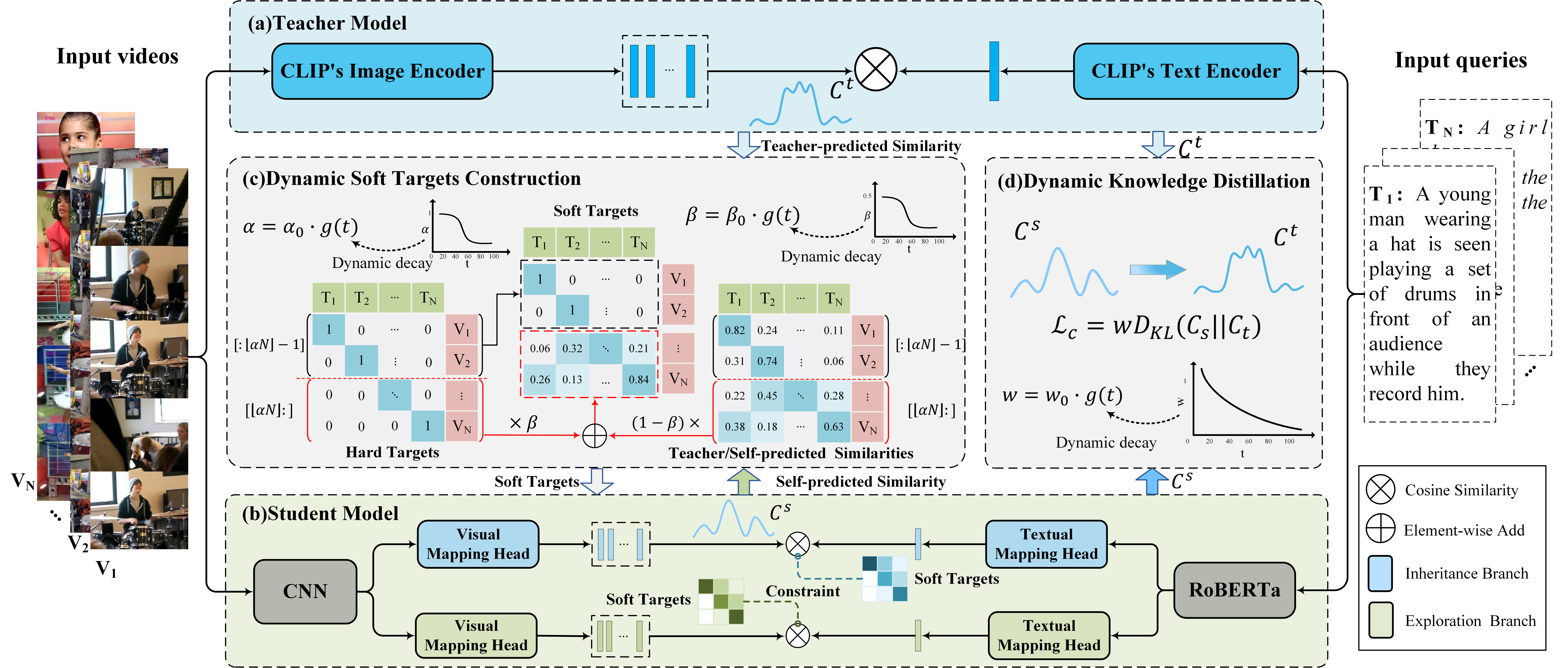}
	\caption{An overview of our proposed DL-DKD++ framework. 
 Given an untrimmed video and a query input, we design a teacher-student network to transfer the knowledge of the large-scale pre-training model CLIP to the PRVR task. In detail, the CLIP model serves as the teacher to provide generalized knowledge. A dual-branch student model is then devised to control the learning effect on knowledge transfer with a dynamic knowledge distillation strategy.
    In particular, an inheritance student branch is introduced to absorb the beneficial knowledge of the teacher model, while an exploration student branch is utilized to explore the task-specific property of the training data. Additionally, a dynamic soft targets construction module is introduced to convert the hard targets used in the conventional InfoNCE loss into soft targets. This allows the model to learn more reasonable partial relevance across different video-text samples.}
	\label{method framework overview}
\vspace{-4mm}
\end{figure*}

\section{Method}
\noindent
\subsection{Overview}
We propose a distillation learning based framework DL-DKD++ for the PRVR task, which mainly consists of two parts as shown in Fig.\ref{method framework overview}: 1) \textit{Teacher-Student dual learning}: 
To leverage the generalization capability of large-scale vision-language pre-trained models, we adopt the CLIP~\cite{radford2021learning} model as the teacher model, which provides semantic guidance for the student model. However, the performance of pre-trained models often varies significantly across different datasets due to domain gaps specific to the different tasks, making it challenging for a single-branch student model to fully benefit from the teacher.
To address this, we design a dual-branch student network composed of an inheritance branch and an exploration branch. The inheritance branch is responsible for absorbing transferable and reliable knowledge from the teacher model, while the exploration branch focuses on learning domain-specific characteristics that may not be well captured by the teacher. These two branches work collaboratively to balance generalization and specialization in learning. Besides, we propose a dynamic knowledge distillation strategy  to adaptively adjusts the reliance on the teacher’s knowledge, helping the overall architecture achieve optimal performance.
2): \textit{Soft Alignment}: To address the limitation of hard binary supervision in existing PRVR methods, we introduce a soft target construction that transforms hard targets into soft targets, enabling the model to better capture the partial relevance between untrimmed videos and textual queries. 
Moreover, to ensure stable and effective learning, we also introduce a dynamic soft target learning strategy. In the early stages of training, we emphasize hard targets to provide clear supervision. As training progresses, the model gradually shifts towards learning from soft targets, 
encouraging the model to learn more flexible and fine-grained semantic alignments.

\subsection{Teacher-Student Framework}
\textbf{CLIP Teacher Model.}
The large-scale vision-language model CLIP~\cite{radford2021learning} was pre-trained on a great amount of image-text data, and is now commonly employed as a strong vision-language backbone enabling zero-shot knowledge transfer to various downstream tasks~\cite{liu2022transfer,wu2022contextual}. 
Therefore, we also resort to CLIP, utilizing it as the teacher model to appropriately guide our student model training. 
\red{Note that other vision-language pre-training models, such as BLIP-2 \cite{li2023blip}, can also be employed here. We conduct experiments with different teacher models in Section \ref{ssec:multi-teacher}, showing the remarkable generalization ability of our proposed framework.}

As depicted in Fig.~\ref{method framework overview}(a), given a video-text pair $(V, Q)$ consisting of a video $V=\{I_i\}_{i=1}^{k}$ of $k$ frames and a textual query $Q$ as input,
we feed them into the CLIP's image and text encoders to obtain the corresponding video feature ${F^{t}}=\{f^{t}_i\}_{i=1}^k\in\mathbb{R}^{d\times k}$ and query feature ${q^{t}}\in\mathbb{R}^{d}$, respectively. 
The video feature is comprised of a sequence of $k$ frame features, and the dimensions of the frame and the query features are both $d$.
Considering that semantic-aligned similarity matters a lot in our retrieval task, we aim to transfer the collected knowledge of video-query semantic-aware similarity distribution from the teacher model to the student model.
Formally, for a pair of video $V$ and query $Q$, their semantic similarity distribution ${C^t}\in\mathbb{R}^k$ is obtained as:
\begin{equation}
\begin{aligned}
    {C^t} = [cos({f^{t}_1}, {q^t}), cos({f^{t}_2}, {q^t}), ..., cos({f^{t}_k, {q^t}})],
\end{aligned}
\end{equation}
where $cos$ denotes the cosine similarity.

Different from many works~\cite{ma2022open,he2021enhancing} devoting to distilling image features from a teacher model to a student model, our teacher model is committed to guide the student model by constraining the consistent semantic similarity distributions between the teacher-student models.

\textbf{Dual-branch Student Model.}
To effectively inherit the teacher model's beneficial knowledge while adapting to domain-specific characteristics present in the training data, we design a dual-branch student model. As illustrated in Fig.~\ref{method framework overview}(b), this student model comprises two complementary branches: an inheritance branch, responsible for knowledge distillation from the teacher, and an exploration branch, dedicated to learning data-specific properties by directly fitting the training data. In the following, we first elaborate on the exploration branch, which aims to mitigate the performance drop caused by domain gaps. Then, we illustrate how we transfer the teacher's knowledge into the inheritance branch.

\subsection{Exploration Student Branch} \label{ssec:ex-loss}
The exploration branch focuses on learning domain-specific characteristics for the PRVR tasks. Here, we carefully illustrate our proposed PRVR in-domain learning with soft targets.

\textbf{Multi-Modal Encoding.} \label{ssec:ex-loss} 
Given an input video $V=\{I_i\}_{i=1}^{k}$, a pre-trained 2D CNN with an $FC$ layer is employed to extract the higher-level CNN features of the
video as ${F^{e'}}\in\mathbb{R}^{z\times k}$,
where each video frame is represented as a $z$-dimensional feature vector. 
Then, after an operation of a standard Transformer with positional embedding and another $FC$ layer, ${F^{e'}}$ is projected into the joint latent space for the latter multi-modal similarity measurement. This encoded visual feature ${F^{e}}\in\mathbb{R}^{z\times k}$ is denoted as:
\begin{equation}
\begin{aligned}
        F^{e} = \{{f^{e}_1}, {f^{e}_2}, ..., {f^{e}_k}\} = FC(Trans(FC(F^{e'})+PE)),
\end{aligned}
\end{equation}
where $PE$ stands for positional embedding, and $Trans$ is a standard Transformer.

For an input query $Q$, following \cite{lei2020tvr,dong2022prvr}, we utilize the pre-trained RoBERTa \cite{liu2019roberta} with an $FC$ layer to generate the word-level features ${Q^{e'}}=\{w^{e'}_i\}_{i=1}^{n_s}\in\mathbb{R}^{z\times {n_s}}$.
To further obtain the contextual features of the query text, we first feed ${Q^{e'}}$ into a standard Transformer to obtain ${Q^{e}}=\{{w^e_i}\}_{i=1}^{n_s}\in\mathbb{R}^{z\times {n_s}}$, and then employ an attention layer to generate the sentence-level feature ${q^{e}}\in\mathbb{R}^{z}$ via attentive aggregation as:
\begin{equation}
\begin{aligned}
    {q^{e}}=\sum_{i=1}^{n_e} \alpha_{i}\times{{w}_i^e},\quad
    \alpha=Softmax({W}  {Q^e} ),
\end{aligned}
\end{equation}
where $Softmax$ denotes the softmax layer, $W\in\mathbb{R}^{ {1} \times z}$ is a trainable variable, and $\alpha \in\mathbb{R}^{{1 \times n_s} } $ indicates the attention vector. ${q^{e}}$ is in the joint latent space with $F^{e}$. 

\textbf{Partial Similarity Computation.} 
In the PRVR task, a video is considered partially relevant to a given textual query as long as it contains query-related content, even though other portions of the video may be irrelevant.
This partial alignment presents a unique challenge, as the similarity computation must effectively capture relevant content while being robust to irrelevant content.
Following the previous work~\cite{dong2022prvr}, we adopt a simple but effective max pooling strategy to compute the similarity. Max pooling allows the model to focus on the most relevant content, even if irrelevant content exists.
With both video and query encoded into the joint latent space, the partial similarity between the input video $V$ and query $Q$ is then measured as:
\begin{equation}
\begin{aligned}
     S(v, q) = \mathop{\max}(\{cos(f^{e}_1, q^{e}), ...,cos(f^{e}_k, q^{e})\}),\\
\end{aligned}
\end{equation}
where $cos(\cdot)$ denotes the cosine similarity function, $max(\cdot)$ denotes the max-pooling operation.

\textbf{Dynamic Soft Targets Construction.}
To learn the matching correspondences between the video and the query, previous PRVR methods typically use a hard binary strategy that labels a video as positive only if it contains content exactly related to the query. 
However, this ignores the specific degree of relevance between the video and the query. 
Therefore, instead of using hard binary labels for supervision, we propose to learn video-text partial relevance with soft labels, as illustrated in Fig.~\ref{method framework overview}(c). Our method constructs soft targets by dynamically combining information from two complementary perspectives: (i) \textit{manually annotated hard targets}, which provide explicit video-to-text alignments, and (ii) \textit{model-predicted similarity scores}, which reflect soft and flexible partial alignments between video and text.
To merge the hard and soft targets,
we not only control the sample-wise proportion of hard-supervised versus soft-supervised pairs, but also integrate score-level soft supervision by interpolating between binary labels and model predictions. 
During training, both the ratio of hard-supervised samples and the weight of binary supervision are gradually reduced, allowing the model to first focus on reliable human-labeled binary signals and then adapt to more flexible and fine-grained soft alignments as training progresses.
In the following, we first illustrate how we balance the hard and soft targets, then we provide the details on how we refine the soft targets into more reliable ones during the pseudo-label learning process.

\subsubsection{Sample-wise Dynamic Target Selection}
To balance hard and soft supervision during training, we introduce a sample-wise dynamic target selection strategy. It progressively selects a portion of samples within each mini-batch to be trained with soft targets, while the remaining samples continue to be trained with hard labels.
Specifically, within a mini-batch of $N$ video-text pairs with manually annotated alignments, the hard targets are represented as a binary identity matrix $I \in \mathbb{R}^{N \times N}$, where $I_{ij} = 1$ if video $v_i$ is the annotated ground truth for text $q_j$, and $0$ otherwise. 
We introduce a dynamic boundary controlled by a parameter $\alpha \in [0,1]$ which determines the proportion of samples trained with hard targets:
\begin{equation}
    I_r=I[:\lfloor\alpha{N}\rfloor-1, :].
\end{equation}
The remaining ones are trained with soft targets. Further,
the parameter $\alpha$ is decayed progressively during training:
\begin{equation}
    \alpha={\alpha}_0\cdot g(t),
\end{equation}
where $\alpha_0$ is an initial weight, $t$ denotes the $t$-th training epoch, and $g(t)$ is a predefined decay function. 
This schedule allows for progressively increasing the partition of soft targets.

\subsubsection{Score-wise Dynamic Target Refinement}
For the soft-supervised target $I_s$, we construct it at the score level by integrating model-predicted soft alignment similarities.
We firstly obtain the exploration branch's self-estimated matrix $T \in \mathbb{R}^{N \times N}$, where each element $T_{ij}$ represents the predicted semantic relevance between video $v_i$ and text $q_j$:
\begin{equation}
    T=\{S(v_i,q_j), 1{\leq}i,j{\leq}N\},
\end{equation}
Then, we combine the binary labels and the corresponding model-predicted similarity scores to obtain the soft targets as:
\begin{equation}\label{eq:soft target construct}
    I_s=\beta\cdot I[\lfloor\alpha{N}\rfloor:, :]+(1-\beta)\cdot T[\lfloor\alpha{N}\rfloor:,:],
\end{equation} 
where $\beta \in [0,1]$ is a dynamic weight controlling the contribution between hard labels and predicted similarity scores in the final soft targets. Similar to $\alpha$, the weight $\beta$ also decays over training time:
\begin{equation}
    \beta={\beta}_0\cdot g(t),
\end{equation}
allowing the model to smoothly evolve from binary alignment constraints to soft relevance modeling, capturing richer supervision signals for partial matches.
\newline
\indent
The full soft targets for text to video alignment are the concatenation of $I_r$ and $I_s$, denoted as:
\begin{equation}
    \widetilde{I}^{t2v}=[I_r|I_s].
\end{equation}
Similarly, the soft targets for video to text alignment $\widetilde{I}^{v2t}$ is also constructed using the transposed guidance similarity matrix ${T}^t$. The resulting soft targets $\widetilde{I}^{t2v}$ and $\widetilde{I}^{t2v}$ are employed in the subsequent improved InfoNCE loss.

\textbf{Training with Soft Targets.}
Following prior works\cite{dong2022prvr}, we employ a combination of triplet ranking loss\cite{faghri2017vse++,dong2021dual} and InfoNCE loss\cite{miech2020end,zhang2021video}  to effectively optimize the exploration branch. While we keep the standard triplet loss to enforce relative distance between positive and negative pairs, we enhance the standard InfoNCE loss by introducing soft targets to better capture partial relevance in PRVR.
Triplet loss focuses on relative similarity, ensuring that a positive pair is closer than negative ones by a margin. In contrast, InfoNCE optimizes absolute similarity scores across all samples. Given this, we apply soft targets only to InfoNCE, where fine-grained supervision on absolute similarity is more impactful.
Formally, given a video-query pair $(v_i, q_j)$ from the mini-batch $\mathcal{B}$, the improved InfoNCE loss with soft targets is defined as:
\begin{equation}
\begin{aligned}\label{eq: soft nce}
    \mathcal{L}_{nce} = 
    -\frac1N\sum_{i=1}^N\sum_{j=1}^N\widetilde{I}^{v2t}_{ij}\log\frac{S(v_i,q_j)}{\sum_{k=1}^N S(v_i,q_k)} \\
    -\frac1N\sum_{i=1}^N\sum_{j=1}^N\widetilde{I}^{t2v}_{ij}\log\frac{S(q_i,v_j)}{\sum_{k=1}^N S(q_i,v_k)},
\end{aligned}
\end{equation}
where $N$ is the size of a mini-batch.
In the standard InfoNCE loss\cite{miech2020end,zhang2021video}, both $\widetilde{I}^{v2t}$ and $\widetilde{I}^{t2v}$ are identity  matrices, where $\widetilde{I}_{ij}=1$ when $i=j$, and $\widetilde{I}_{ij}=0$ otherwise. This means it treats all positives equally with binary labels, which is suboptimal for PRVR where video-text pairs may exhibit varying degrees of partial relevance. To address this, we propose converting the hard targets into soft targets, enabling the model to transition from rigid binary alignment to more flexible and relevance-degree aware matching.
In parallel, triplet loss guides the model to maintain robust global ranking. Given a positive video-query pair $(v, q)$, the triplet ranking loss over the mini-batch $\mathcal{B}$ is defined as:
\begin{equation}
\begin{split}
    \mathcal{L}_{trip}=\frac{1}{N}\sum_{(q,v) \in \mathcal{B}}[max(0,m+{S}(q^-, v)-{S}(q, v))\\
    +max(0,m+{S}(q, v^-)-{S}( q,v))],
\end{split}
\end{equation}
where $m$ is the margin constant.
Besides, $q^-$ and $v^-$ respectively indicate a negative sentence sample for $v$ and a negative video sample for $s$.

Finally, the overall similarity learning loss of the exploration branch $\mathcal{L}_E$ is defined as:
\begin{equation}
    \mathcal{L}_E=\mathcal{L}_{nce}+\mathcal{L}_{trip}.
\end{equation}

\subsection{Inheritance Student Branch} \label{ssec:ex-loss}
The inheritance branch is expected to learn the collected knowledge from the semantic similarity distribution ${C^t}$ of the teacher model. As illustrated in Fig.~\ref{method framework overview}(b), the inheritance branch shares the same network architecture as the exploration branch. Given an input video $V$ and a query $Q$, they are encoded into frame-level features ${F^{s}}\in\mathbb{R}^{z\times k} = \{{f^{s}_1}, {f^{s}_2}, ..., {f^{s}_k}\}$ and a sentence-level feature ${q^{s}\in\mathbb{R}^{z}}$ in the same manner, respectively.

To absorb knowledge from the teacher by learning the consistency of video-query semantic similarity distribution between the teacher and student branches further, we first calculate the similarity distribution ${C^s}\in\mathbb{R}^k$ of current student branch between ${F^{s}}$ and ${q^{s}}$:
\begin{equation}
\begin{aligned}
    {C^s}=[cos({f^{s}_1},{q^{s}}), cos({f^{s}_2},{q^{s}}),...,cos({f^{s}_k},{q^{s}})].
\end{aligned}
\end{equation}
Then, we design a distribution distillation to transfer knowledge from the pre-trained teacher model to the inheritance student branch.  
Specifically, our distribution distillation strategy is to capture the consistency of the similarity distributions of the teacher and the student. A similarity consistency constraint is configured to guide the learning of the inheritance branch with the teacher model\cite{wu2022contextual}.
In detail, given the teacher-similarity distribution ${C^t}$ and the student-similarity distribution ${C^s}$ of a video-text pair $(V, Q)$, the semantic consistency loss ${\mathcal{L}_c}$ is formulated by exploiting the KL divergence as:
\begin{equation}
\begin{aligned}
    {\mathcal{L}_c}={D_{KL}}({C^{s}}||{C^{t}})
    =\sum_{i=1}\limits^{k}     {{C^{s}_i} log{\frac{{C^{s}_i}}{{C^{t}_i}}}},
\end{aligned}
\end{equation}
where the subscript $i$ indicates the $i$-th elements in the corresponding similarity distribution.

Besides the knowledge distillation loss, we also employ triplet loss $\mathcal{L}_{trip}^{s}$ and improved InfoNCE loss $\mathcal{L}_{{nce}}^{s}$ for self-similarity learning in this branch.
In the inheritance branch, the model-predicted similarity matrix used for soft target construction  (Eq.~\ref{eq:soft target construct}) is derived from the CLIP teacher model.
Overall, to train our inheritance student branch, we simultaneously learn the similarity consistency and the self-similarity. The final loss $\mathcal{L}_I$ of this branch is termed as:
\begin{equation}\label{eq: distill loss}
\begin{aligned}
    \mathcal{} \mathcal{L}_I = {w}{{\mathcal{L}_c}} + \mathcal{L}_{{nce}}^{s} + \mathcal{L}_{trip}^{s},
\end{aligned}
\end{equation}
where $w$ is a hyper-parameter for loss balance.

\subsection{Dynamic Knowledge Distillation} \label{ssec:ex-loss}
Although we can directly jointly learn the two student branches, this training process remains two-aspect concerns:
Firstly, as we mentioned, the CLIP teacher model may have huge performance differences on different datasets due to their task-specific domain gaps. Therefore, when the teacher model is of mediocre performance, how to reduce the impact of the inheritance branch while strengthening the exploration branch learning is important.
Secondly, it is worth noticing that continuously pushing the student model to mimic the similarity distribution of the teacher model during the whole training period may limit the student model in acquiring data-specific knowledge. 

Based on the above two observations, we propose a dynamical learning paradigm to adjust and distill the knowledge learned from the dual branches.
The main idea is:
\textit{learning more knowledge from the teacher at the beginning of the training when the knowledge of the teacher is beneficial while learning more from the on-site data gradually otherwise when the student model gets stronger.} 
Specifically, to obtain a more balanced and better distillation result from the dual-branch learning, a dynamic distillation strategy is introduced.
It is devised to tune the hyper-parameter $w$ in Eq.(\ref{eq: distill loss}) online during the model training instead of setting it to a fixed constant one like most previous works. 
At the beginning of the training, we set a larger initial value to $w$ to learn more knowledge from the teacher, then we decay $w$ smoothly according to the training epochs. Formally, $w$ is computed as:
\begin{equation}
\begin{aligned}
\label{L_kl_loss}
    w = {w_0}{\cdot}g(t),
\end{aligned}
\end{equation}
where ${w_0}$ is an initial weight, $t$ indicates $t$-th epoch during the training, and $g(\cdot)$ is the decay strategy function.

\subsection{Overall Training and Inference}
During the entire training process, we exclusively optimize the student model by jointly minimizing the losses from both the inheritance and exploration branches. 
The overall training loss $\mathcal{L}$ is formulated as:

\begin{equation}
\begin{aligned}
    \mathcal{L} = {\mathcal{L}_I}+{{\mathcal{L}_E}}.
\end{aligned}
\end{equation}

During the inference, only the student model is utilized to obtain the retrieval results.
Given a video-text pair, we first calculate their similarities from both inheritance and exploration branches, resulting in $S_I(Q,V)$ and $S_E(Q,V)$.
The final similarity is computed as:
\begin{equation}\label{eq:beta}
    S(Q,V)= (1-\sigma) S_I(Q,V) + \sigma S_E(Q,V),
\end{equation}
where $\sigma$ is a hyper-parameter to balance the two similarities.
Given a textual query, all candidate videos are sorted in terms of their final similarities with the query.

\section{Experiments} \label{sec:eval}

\subsection{Experimental Setup} \label{ssec:exp-set}

\subsubsection{Datasets} 

\blue{
In order to validate the effectiveness of our model, we adopt the long untrimmed video datasets ActivityNet Captions \cite{krishna2017dense}, TVR \cite{lei2020tvr}, and Charades-STA \cite{gao2017tall} to implement experiments. 
Note that the pre-trained CLIP performs well on Activitynet Captions, but mediocrely on TVR and Charades-STA. Here we briefly introduce these three datasets in the following.
}

\textit{ActivityNet-Captions} \cite{krishna2017dense}: ActivityNet-Captions is originally developed for dense video captioning task. As captions are partially relevant with the corresponding
videos (a caption is typically associated with a specific moment in
a video), it has been re-purposed for partially relevant video retrieval.
It contains around 20K videos from YouTube, and the average length of videos is around 118 seconds. 
On average, each video has around 3.7 moments with a corresponding sentence description.
For a fair comparison, we adopt the same data partition used in \cite{dong2022prvr}.
For ease of reference, we refer to the dataset as ActivityNet.

\blue{
\textit{TVshow Retrieval (TVR)} \cite{lei2020tvr}: TVR is originally developed for video corpus moment retrieval, and now can be also used for partially relevant video retrieval.
It contains 21.8K videos collected from 6 TV shows, and the average length of videos is around 76 seconds.
Each video is associated with 5 natural language sentences that describe a specific moment in the video. As a moment is typically a part of a video, sentences are partially relevant to videos.
We utilize the same data partition in \cite{zhang2020hierarchical,zhang2021video,dong2022prvr}.
}

\textit{Charades-STA} \cite{gao2017tall}: This dataset contains 6,670 videos with 16,128 sentence descriptions, with an average of approximately 2.4 moments and their corresponding sentence descriptions per video. The official data partition is employed for the purposes of model training and evaluation.

\begin{figure}[tb!]
\centering
\includegraphics[width=0.7\columnwidth]{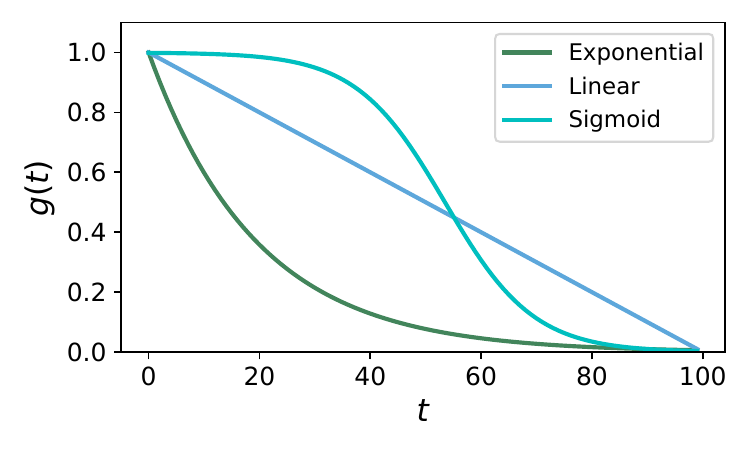}
\vspace{-4mm}
\caption{Different decay strategies during the training.}
\label{fig:decay-strategy-on-train}
\vspace{-2mm}
\end{figure}

\subsubsection{Evaluation Metrics}
\blue{
Following the previous work~\cite{dong2022prvr}, we utilize
the rank-based metrics, namely $R@K$ ($K = 1, 5, 10, 100$).
$R@K$ stands for the fraction of queries that correctly retrieve desired items
in the top $K$ of the ranking list.
The performance is reported in percentage (\%).
The SumR is also utilized as the overall performance, which is defined as the sum of all recall scores.
Higher scores indicate better performance.
}

\begin{table*}[tb!]
\renewcommand{\arraystretch}{1.6}
\caption{Performance comparison with the state-of-the-art on the three datasets. Models are sorted in ascending order in terms of their overall performance. * indicates the performance reproduced using official open-source code.}
\vspace{-2mm}
\label{tab:sota-all}
\centering 
\scalebox{0.87}{
\centering
\begin{tabular}{@{}ccccccccccccccccccc@{}}
\bottomrule
\hline
\multirow{2}{*}{\textbf{Model}} & \textbf{} & \multicolumn{5}{c}{\textbf{TVR}}                                                & \textbf{} & \multicolumn{5}{c}{\textbf{ActivityNet-Captions}}                                        & \textbf{} & \multicolumn{5}{c}{\textbf{Charades-STA}}                                        \\ \cline{3-7} \cline{9-13} \cline{15-19} 
                                & \textbf{} & \textbf{R@1}  & \textbf{R@5}  & \textbf{R@10} & \textbf{R@100} & \textbf{SumR}  & \textbf{} & \textbf{R@1}  & \textbf{R@5}  & \textbf{R@10} & \textbf{R@100} & \textbf{SumR}  & \textbf{} & \textbf{R@1} & \textbf{R@5} & \textbf{R@10} & \textbf{R@100} & \textbf{SumR} \\ \hline
\multicolumn{19}{l}{\textit{T2VR models:}}                                                                                                                                                                                                                                                                             \\
W2VV\cite{dong2018predicting}                            &           & 2.6           & 5.6           & 7.5           & 20.6           & 36.3           &           & 2.2           & 9.5           & 16.6          & 45.5           & 73.8           &           & 0.5          & 2.9          & 4.7           & 24.5           & 32.6          \\
HTM\cite{miech2019howto100m}                             &           & 3.8           & 12.0          & 19.1          & 63.2           & 98.2           &           & 3.7           & 13.7          & 22.3          & 66.2           & 105.9          &           & 1.2          & 5.4          & 9.2           & 44.2           & 60.0          \\
HGR\cite{chen2020fine}                             &           & 1.7           & 4.9           & 8.3           & 35.2           & 50.1           &           & 4.0           & 15.0          & 24.8          & 63.2           & 107.0          &           & 1.2          & 3.8          & 7.3           & 33.4           & 45.7          \\
RIVRL\cite{dong2022reading}                           &           & 9.4           & 23.4          & 32.2          & 70.6           & 135.6          &           & 5.2           & 18.0          & 28.2          & 66.4           & 117.8          &           & 1.6          & 5.6          & 9.4           & 37.7           & 54.3          \\
VSE++\cite{faghri2017vse++}                           &           & 7.5           & 19.9          & 27.7          & 66.0           & 121.1          &           & 4.9           & 17.7          & 28.2          & 67.1           & 117.9          &           & 0.8          & 3.9          & 7.2           & 31.7           & 43.6          \\
DE++\cite{dong2021dual}                            &           & 8.8           & 21.9          & 30.2          & 67.4           & 128.3          &           & 5.6           & 18.8          & 29.4          & 67.8           & 121.7          &           & 1.7          & 5.6          & 9.6           & 37.1           & 54.1          \\
DE\cite{dong2019dual}                               &           & 7.6           & 20.1          & 28.1          & 67.6           & 123.4          &           & 5.6           & 18.7          & 29.7          & 68.8           & 122.6          &           & 1.5          & 5.7          & 9.5           & 36.9           & 53.7          \\
W2VV++\cite{li2019w2vv++}                          &           & 5.0           & 14.7          & 21.7          & 61.8           & 103.2          &           & 5.4           & 18.7          & 29.7          & 68.8           & 122.6          &           & 0.9          & 3.5          & 6.6           & 34.3           & 45.3          \\
CE\cite{liu2019use}                              &           & 3.7           & 12.8          & 20.1          & 64.5           & 101.1          &           & 5.5           & 19.1          & 29.9          & 71.1           & 125.6          &           & 1.3          & 4.5          & 7.3           & 36.0           & 49.1          \\ \hline
\multicolumn{19}{l}{\textit{VCMR models w/o moment localization:}}                                                                                                                                                                                                                                                     \\
ReLoCLNet\cite{zhang2021video}                       &           & 10.7          & 28.1          & 38.1          & 80.3           & 157.1          &           & 5.7           & 18.9          & 30.0          & 72.0           & 126.6          &           & 1.2          & 5.4          & 10.0          & 45.6           & 62.3          \\
XML\cite{lei2020tvr}                             &           & 10.0          & 26.5          & 37.3          & 81.3           & 155.1          &           & 5.3           & 19.4          & 30.6          & 73.1           & 128.4          &           & 1.6          & 6.0          & 10.1          & 46.9           & 64.6          \\ \hline
\multicolumn{19}{l}{\textit{PRVR models:}}                                                                                                                                                                                                                                                                             \\
UMT-L\cite{li2023unmasked}                           &           & 13.7          & 32.3          & 43.7          & 83.7           & 173.4          &           & 6.9           & 22.6          & 35.1          & 76.2           & 140.8          &           & 1.9          & \textbf{7.4} &  12.1    & 48.2           & 69.6          \\
MS-SL\cite{dong2022prvr}                           &           & 13.5          & 32.1          & 43.4          & 83.4           & 172.4          &           & 7.1           & 22.5          & 34.7          & 73.1           & 140.1          &           & 1.8          & 7.1          & 11.8          & 47.7           & 68.4          \\
JSG*\cite{chen2023joint}                             &           & 11.3          & 29.1          & 39.6          & 80.9           & 161.0          &           & 6.7           & 22.5          & 34.8          & 76.2           & 140.3          &           & 1.8          & 7.2          & 11.9          & 48.3           & 69.2          \\
GMMFormer*\cite{wang2024gmmformer}                       &           & 13.5          & 33.1          & 44.4          & 84.3           & 175.3          &           & 7.5           & 24.1          & 36.0          & 75.5           & 143.1          &           & 1.8          &  7.3    & 12.0          & \textbf{50.0}  & 71.0    \\
DL-DKD++(Ours)                  &           & \textbf{15.3} & \textbf{36.0} & \textbf{47.5} & \textbf{86.0}  & \textbf{184.8} &           & \textbf{8.3} & \textbf{25.5} & \textbf{38.3} & \textbf{77.8}  & \textbf{149.9} &           & \textbf{1.9} & 7.1          & \textbf{12.3} & 49.8     & \textbf{71.1} \\
\bottomrule
\hline
\end{tabular}
}
\end{table*}
\subsubsection{Implementation Details}
For the CLIP teacher model, we adopt a Vision Transformer based ViT-B/32 provided by OpenAI, and encode video frames and query sentences to 512-D features.
For the student model, we directly utilize the video and sentence features provided by \cite{dong2022prvr} as the input. The dimensional sizes of the video features extracted by the pre-trained CNN model are 1024 for the ActivityNet-Captions and Charades-STA datasets, and 3072 for the TVR datasets. The dimensions of all the above features are linearly reduced to 384 further for the convenience of the Transformer's (384 hidden sizes, 4 attention heads) feature encoding.
For a textual query, we employ the pre-trained Roberta \cite{liu2019roberta} to extract a feature with 1024 dimension firstly, reduce the dimension of the feature to 384 further, and then feed the feature to a Transformer (384 hidden sizes, 4 attention heads) for feature encoding.
For the model training, following \cite{dong2018predicting}, we use the early stop schedule that the model will stop when the evaluated SumR exceeds 10 epochs without promotion. The maximum number of epochs is set to 100.

During the inference, we empirically set the weights of the inheritance branch and the exploration branch to 0.3 and 0.7 for similarity fusion.
Additionally, we use PyTorch to build the model framework and train models on NVIDIA RTX 3090 GPU with a batch size of 128.

For the decay strategy, inspired by prior work on scheduled learning \cite{bengio2015scheduled, liu2021scheduled}, we explore three types of decay functions to gradually adjust the corresponding weights during training.
As illustrated in Fig.~\ref{fig:decay-strategy-on-train}, the three decay strategy functions are:
\begin{itemize}
    \item Exponential decay: $g(t)={k^t}$, where $k$ $<$ 1 is the factor that control the decay trend.
    \item Linear decay: $g(t)=kt+b$, where $k$ $<$ 0 and $b$ is for controlling the downward trend of the slash.
    \item Sigmoid decay: $g(t)=\frac{k}{k+{e^{\frac{t}{k}}}}$, where $k$ is a hyper-parameter to control the decay.
\end{itemize}
In our implementation, we apply exponential decay with an initial value of 0.1 and a decay factor of $k = 0.95$ for $w$. For $\alpha$ and $\beta$, we initialize both to 0.8 and adopt a sigmoid decay function with $k = 800$. Experimental results demonstrate that all three decay strategies perform consistently well, highlighting the robustness of our training framework.

\begin{figure}[tb!]
    \centering
    \subfloat[\textrm{TVR}]{
    \includegraphics[width=0.95\columnwidth]{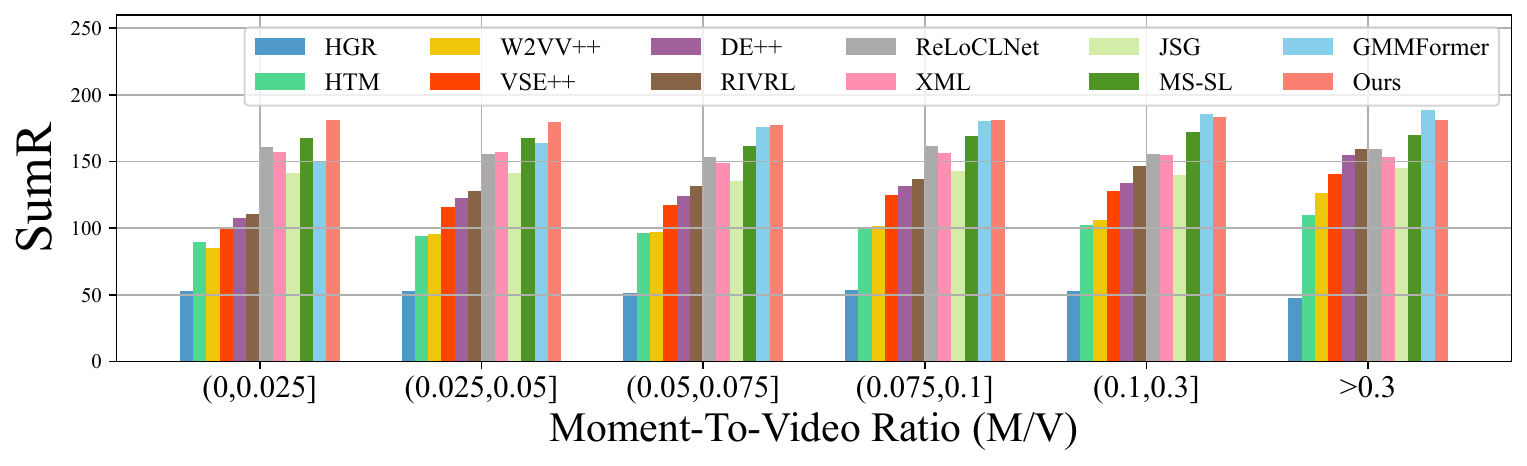}
    }
    \\[-0.5mm]
    \subfloat[\textrm{ActivityNet}]{
    \includegraphics[width=0.95\columnwidth]{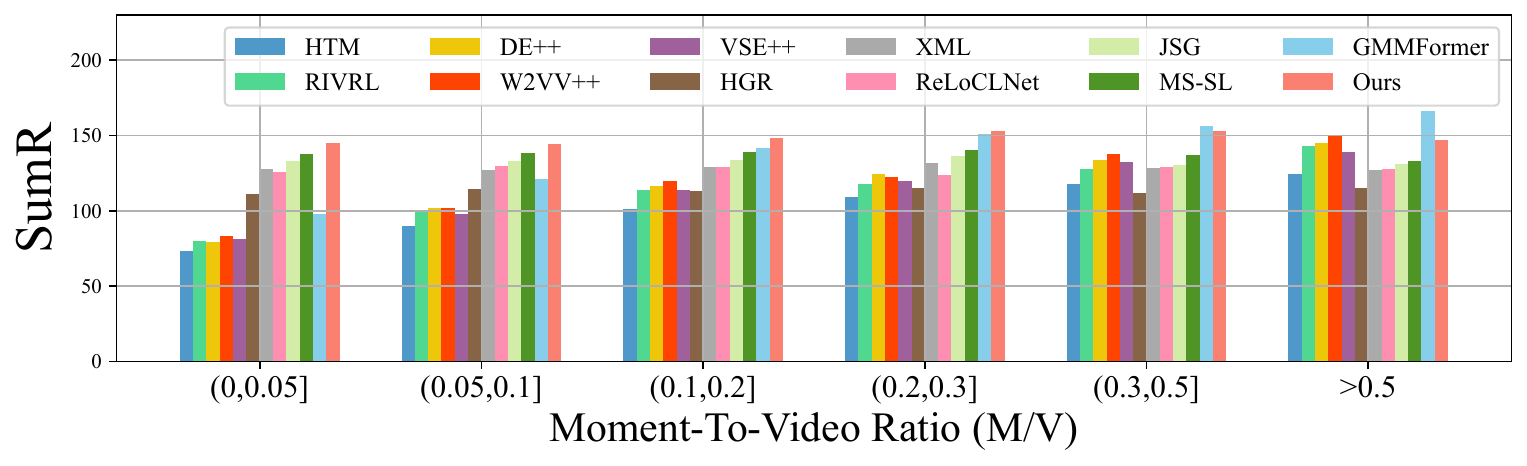}
    }
    \\[-0.5mm]
    \subfloat[\textrm{Charades-STA}]{
    \includegraphics[width=0.95\columnwidth]{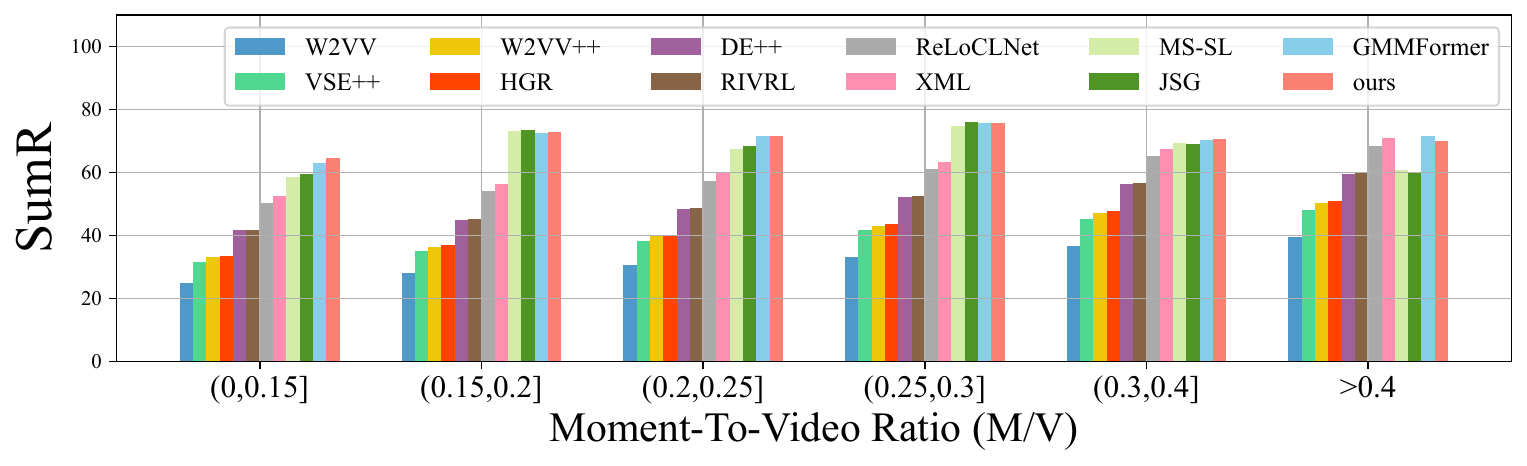}
    }

\vspace{-2mm}
\caption{Performance of different models on various types of queries. Queries are grouped according to their M/V. The smaller M/V indicates more challenging of queries.}
\vspace{-3mm}
\label{fig:group_recalls}
\end{figure}

\subsection{Comparison with the State-of-the-Art} \label{ssec:exp-sota}
Table \ref{tab:sota-all} summarizes performance comparison on three datasets: ActivityNet, TVR, and Charades-STA,
where our proposed model outperforms all the competitor models. 
Among all methods, only our model utilizes the knowledge distillation strategy, and the results justify the viability of using the knowledge of the pre-training vision-language model for partially relevant video retrieval.
In addition, although the previous PRVR models \emph{e.g.,} JSG\cite{chen2023joint} and GMMFormer\cite{wang2024gmmformer}, also utilize the two-branch framework, their two branches solely learn from the training data without extra knowledge. 
By contrast, in our model with the dual learning paradigm, one branch mainly learns from the teacher model and the other learns from the training data. The better performance of our model demonstrates the effectiveness of our proposed dual learning paradigm.
Moreover, unlike other methods, our model uniquely explores and exploits the degree of correlation within positive video-text pairs while identifying latent partially relevant pairs in negative video-text pairs through soft targets. The superior performance of our method further demonstrates that learning soft targets between video-text pairs is more suitable than hard targets for tackling the challenging PRVR task.

\blue{
Thus far, all the comparisons are holistic. To gain a more fine-grained comparison, we group the test queries according to their \textit{moment-to-video ratio} (M/V) \cite{dong2022prvr}.
M/V of the query is defined as its relevant moment’s length ratio in the entire video.
The smaller M/V indicates less relevant content while more irrelevant content in the target video with respect to the query, showing more challenging of the corresponding queries. 
To ensure a fair comparison, we adjust the M/V intervals for each group to maintain a balanced number of samples across groups.
Fig.~\ref{fig:group_recalls} presents the grouped results on all three benchmarks. On ActivityNet and TVR, our model consistently outperforms competing methods across most M/V groups, further validating its effectiveness for the PRVR task. On the challenging Charades-STA dataset, our method achieves performance comparable to GMMFormer.
Although GMMFormer slightly outperforms our model in groups of high M/V, it performs worse in groups of low M/V. This suggests that GMMFormer mainly captures global relevance between text and untrimmed video, but struggles to learn partial relevance.
Moreover, our proposed model achieves more balanced performance across all groups, which shows that our model is less sensitive to irrelevant content in videos.
}

\begin{table*}[tb!]
\renewcommand{\arraystretch}{1.6}
\caption{The effectiveness of dual learning with both inheritance and exploration branches.
Our proposed model not only outperforms the single-branch counterparts but also performs better than simple two-branch baselines.}

\label{tab:effect-of-dynamic-distillation}
\centering 
\scalebox{0.82}{
\begin{tabular}{cccccccccccccccccccc}
\bottomrule
\hline
\multicolumn{2}{c}{\textbf{Branch}}                     & \textbf{} & \multicolumn{5}{c}{\textbf{TVR}}                                                & \textbf{} & \multicolumn{5}{c}{\textbf{ActivityNet-Captions}}                              & \textbf{} & \multicolumn{5}{c}{\textbf{Charades-STA}}                                    \\ \cline{1-2} \cline{4-8} \cline{10-14} \cline{16-20} 
\textbf{Inheritance}       & \textbf{Exploration}       & \textbf{} & \textbf{R@1}  & \textbf{R@5}  & \textbf{R@10} & \textbf{R@100} & \textbf{SumR}  & \textbf{} & \textbf{R@1} & \textbf{R@5}  & \textbf{R@10} & \textbf{R@100} & \textbf{SumR}  & \textbf{} & \textbf{R@1} & \textbf{R@5} & \textbf{R@10} & \textbf{R@100} & \textbf{SumR} \\ \cline{1-14} \cline{16-20} 
\ding{51} & \ding{55} &           & 13.4          & 32.9          & 44.0          & 83.4           & 173.7          &           & 7.7          & 23.9          & 36.4          & 76.5           & 144.5          &           & 1.7          & 5.9          & 10.1          & 45.6           & 63.3          \\
\ding{55} & \ding{51} &           & 13.1          & 32.4          & 43.5          & 83.0           & 172.0          &           & 7.1          & 23.4          & 35.6          & 75.6           & 141.6          &           & 1.6          & 6.0          & 10.0          & 46.1           & 63.8          \\
\ding{51} & \ding{51} &           & \textbf{15.3} & \textbf{36.0} & \textbf{47.5} & \textbf{86.0}  & \textbf{184.8} & \textbf{} & \textbf{8.3} & \textbf{25.5} & \textbf{38.3} & \textbf{77.8}  & \textbf{149.9} &           & \textbf{1.9} & \textbf{7.1} & \textbf{12.3} & \textbf{49.8}  & \textbf{71.1} \\ \hline
\multicolumn{2}{c}{Double-Inheritance}                  &           & 14.2          & 34.5          & 46.4          & 85.0           & 180.1          &           & 7.2          & 24.4          & 37.9          & 77.3           & 146.8          &           & 1.6          & 7.0          & 11.8          & 48.7           & 69.1          \\
\multicolumn{2}{c}{Double-Exploration}                  &           & 14.5          & 34.3          & 46.2          & 85.5           & 180.2          &           & 7.8          & 23.5          & 36.2          & 77.5           & 145.0          &           & 1.7          & 6.9          & 11.5          & 49.4           & 69.5          \\
\bottomrule
\hline
\end{tabular}
}
\end{table*}

\begin{table*}[tb!]
\renewcommand{\arraystretch}{1.6}
\caption{The effectiveness of dynamic knowledge distillation.
Note that \textit{Fixed} indicates the model using knowledge distillation with a fixed weight during the training.
}

\label{tab:total-performance}
\centering 
\scalebox{0.83}{
\begin{tabular}{cccccccccccccccccccc}
\bottomrule
\hline
\multirow{2}{*}{\textbf{Branch}} & \multirow{2}{*}{\textbf{Distillation}} & \textbf{} & \multicolumn{5}{c}{\textbf{TVR}}                                                & \textbf{} & \multicolumn{5}{c}{\textbf{ActivityNet-Captions}}                              & \textbf{} & \multicolumn{5}{c}{\textbf{Charades-STA}}                                    \\ \cline{4-8} \cline{10-14} \cline{16-20} 
                                 &                                        & \textbf{} & \textbf{R@1}  & \textbf{R@5}  & \textbf{R@10} & \textbf{R@100} & \textbf{SumR}  & \textbf{} & \textbf{R@1} & \textbf{R@5}  & \textbf{R@10} & \textbf{R@100} & \textbf{SumR}  & \textbf{} & \textbf{R@1} & \textbf{R@5} & \textbf{R@10} & \textbf{R@100} & \textbf{SumR} \\ \hline
\multirow{3}{*}{\textbf{Single}} & \ding{55}                                      &           & 13.1          & 32.4          & 43.5          & 83.0           & 172.0          &           & 7.1          & 23.4          & 35.6          & 75.6           & 141.6          &           & 1.6          & 6.0          & 10.0          & \textbf{46.1}  & 63.8          \\
                                 & \ding{51}(Fixed)                               &           & 11.0          & 28.5          & 38.8          & 80.0           & 158.3          &           & 7.3          & 23.7          & 35.8          & 75.8           & 142.7          &           & 1.5          & 6.0          & 10.4          & 45.5           & 63.4          \\
                                 & \ding{51}(Dynamic)                             &           & \textbf{13.4} & \textbf{32.9} & \textbf{44.0} & \textbf{83.4}  & \textbf{173.7} &           & \textbf{7.7} & \textbf{23.9} & \textbf{36.4} & \textbf{76.5}  & \textbf{144.5} &           & \textbf{1.7} & \textbf{6.2} & \textbf{10.6} & 45.6           & \textbf{64.1} \\ \hline
\multirow{3}{*}{\textbf{Dual}}   & \ding{55}                                      &           & 14.5          & 34.3          & 46.2          & 85.5           & 180.2          &           & 7.8          & 23.5          & 36.2          & 77.5           & 145.0          &           & 1.7          & 6.9          & 11.5          & 49.4           & 69.5          \\
                                 & \ding{51}(Fixed)                               &           & 14.6          & 34.6          & 46.2          & 85.7           & 181.1          &           & 7.7          & 24.9          & 36.7          & 77.1           & 146.5          &           & 1.8          & 7.3          & 11.7          & 49.0           & 69.8          \\
                                 & \ding{51}(Dynamic)                             &           & \textbf{15.3} & \textbf{36.0} & \textbf{47.5} & \textbf{86.0}  & \textbf{184.8} & \textbf{} & \textbf{8.3} & \textbf{25.5} & \textbf{38.3} & \textbf{77.8}  & \textbf{149.9} &           & \textbf{1.9} & \textbf{7.1} & \textbf{12.3} & \textbf{49.8}  & \textbf{71.1} \\             
\bottomrule                                 
\hline
\end{tabular}}
\end{table*}

\subsection{Investigation on Two-Branch Structure} \label{ssec:ablation}
\subsubsection{Effectiveness of the Dual Learning Paradigm}
In order to verify the effectiveness of our proposed dual learning with both inheritance and exploration branches, we first compare it to the counterparts with the inheritance branch or exploration branch only.
As shown in Table \ref{tab:effect-of-dynamic-distillation}, the model integrating both branches consistently performs best across all three datasets, demonstrating the effectiveness of our proposed dual learning architecture. 
Additionally, we also try to verify whether the improvements come from the combination of two branches. We compare our model to the baselines of simply combining two exploration branches (Dual-exploration) or two inheritance branches (Dual-inheritance) without our dynamic distillation strategy. Their worse performance compared to ours demonstrates that the architecture of our dual-branch exploration and inheritance with the dynamic distillation contributes a lot to the final performance.

\begin{figure}[t!]
        \centering
        \includegraphics[width=0.7\linewidth]{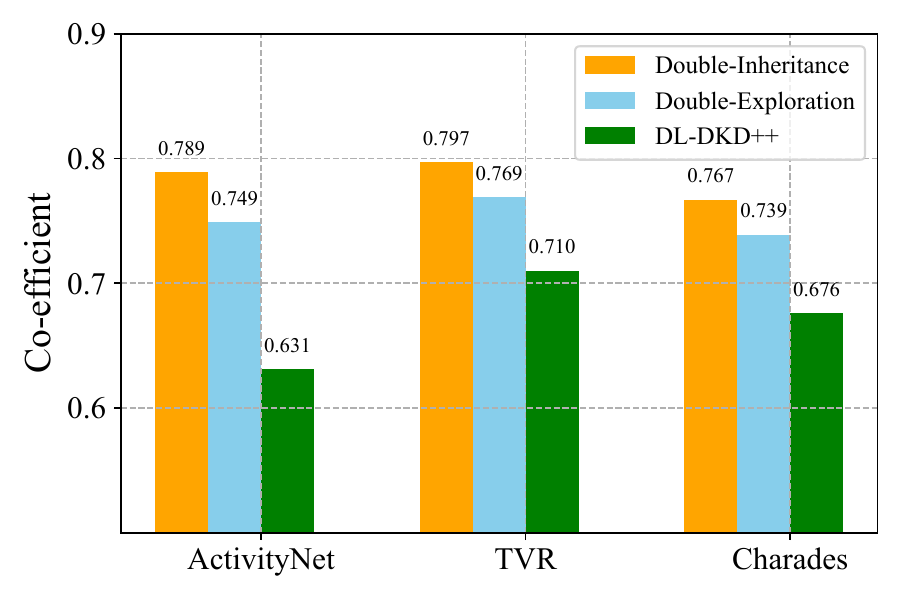}
        \caption{
        Complementarity between the two branches across three datasets. A lower correlation coefficient indicates that the two branches are more complementary. 
	}
	\label{fig:Complementarity}

\end{figure}

\subsubsection{Complementarity between the Two Branches}
Recall that the inheritance branch and the exploration branch in our model are designed to capture distinct aspects of knowledge. To validate this design, we evaluate the complementarity between the two branches by computing the Pearson correlation coefficient between their similarity distributions, \ie $C^t$ and $C^s$, respectively.
For comparison, we also calculate the correlation coefficients for common two-branch baselines (\ie Double-Inheritance and Double-Exploration). As shown in Fig.~\ref{fig:Complementarity}, our model consistently yields lower correlation scores than the baselines across all three datasets.
A lower Pearson correlation coefficient indicates that the two branches produce less similar outputs, suggesting they learn more distinct and complementary representations. These results support the effectiveness of our dual-branch design and help explain why our model outperforms the two-branch baselines.

\subsection{Investigation on Knowledge Distillation}
\subsubsection{Effectiveness of Dynamic Knowledge Distillation}
Table~\ref{tab:total-performance} shows an ablation study on the effectiveness of dynamic knowledge distillation (KD) for both single-branch and dual-branch network architectures. We compare three variants: models trained without KD, with fixed-weight KD, and with our proposed dynamic KD. The fixed weight is set to 0.1, consistent with the initial weight used in our full model.
Under the single-branch setting, applying fixed-weight KD yields mixed outcomes: it improves performance on ActivityNet-Captions but degrades results on TVR and Charades-STA. 
Recall that the teacher model performs well on ActivityNet-Captions, while mediocre on the other two datasets.
This suggests that fixed-weight KD is suboptimal when the teacher model is not strong. In contrast, our dynamic KD consistently outperforms both the baseline and fixed KD across all three datasets, demonstrating its adaptability to varying teacher quality.
For the dual-branch configuration, our dynamic KD consistently achieves the best overall results, further validating its effectiveness. These findings highlight not only the benefits of dual-branch learning but also the flexibility of dynamic KD over conventional fixed-weight approaches.

\begin{figure*}[htbp]
\centering

\subfloat[TVR]{
    \includegraphics[width=0.31\linewidth]{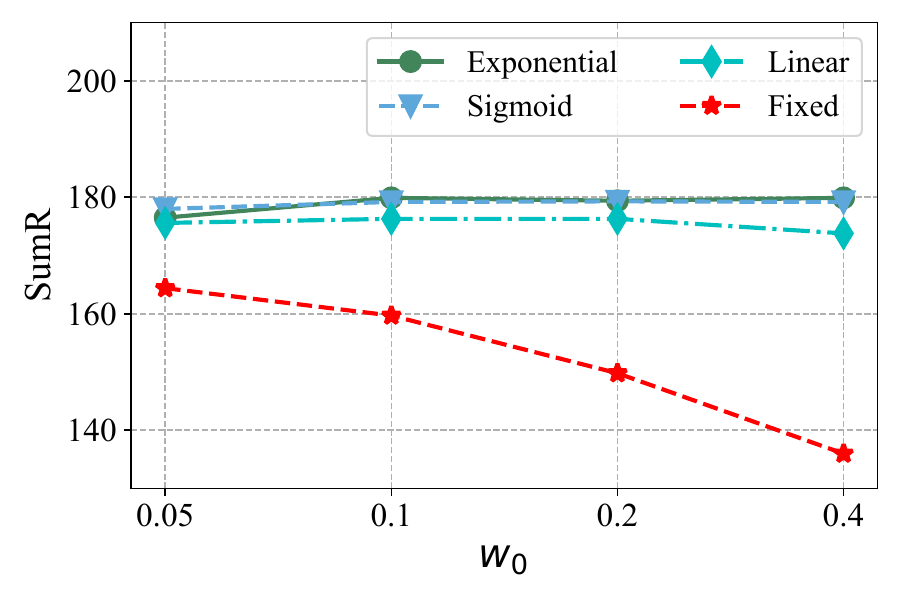}
}
\hfill
\subfloat[ActivityNet Captions]{
    \includegraphics[width=0.31\linewidth]{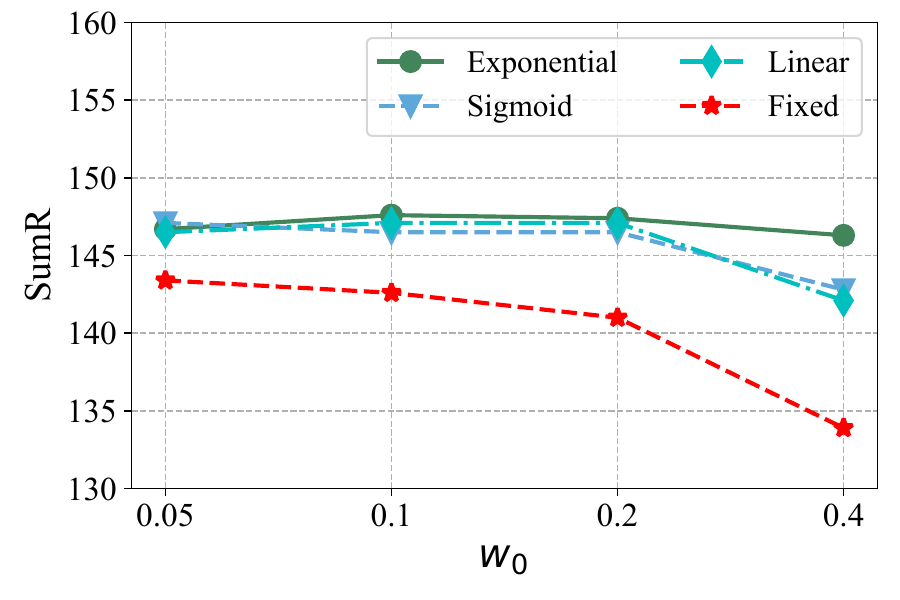}
}
\hfill
\subfloat[Charades-STA]{
    \includegraphics[width=0.31\linewidth]{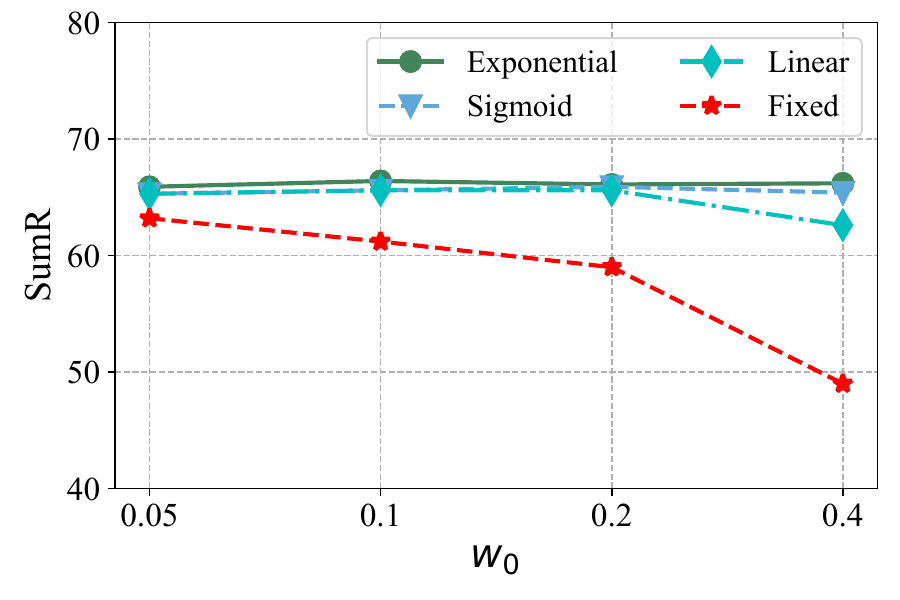}
}

\vspace{1mm}
\caption{The influence of decay strategies in our dynamic knowledge distillation.
The three decay strategies give comparable performances. Besides, they are not very sensitive to the initial weight, making them more robust and easier to tune in practice.}
\vspace{-3mm}
\label{fig:dynamic_distillation_robust}
\end{figure*}

\subsubsection{Influence of Decay Strategies}
We further investigate the impact of different decay strategies for the dynamic knowledge distillation weight $w$. Specifically, we compare three decay schedules, \ie exponential, sigmoid, and linear, across various initial weights. For a more comprehensive analysis, we also include a baseline where $w$ is kept fixed throughout training. The results are presented in Fig.~\ref{fig:dynamic_distillation_robust}.
Across all three datasets, the decay-based strategies exhibit comparable performance, consistently outperforming the fixed-weight baseline. This highlights the robustness of dynamic knowledge distillation to the choice of decay schedule. Among them, exponential decay demonstrates relatively more stable performance across different initial weights, so we adopt it as our default strategy.
Moreover, we observe that dynamic knowledge distillation is significantly less sensitive to the choice of the initial weight $w_0$ compared to the fixed strategy. This robustness greatly reduces the burden of hyper-parameter tuning, making dynamic KD a more practical choice for real-world applications.

\begin{table}[tb!]
\renewcommand{\arraystretch}{1.6}
\caption{\red{Performance with different teacher models. Our proposed framework supports various teacher models, and also allows for distilling for multiple teachers jointly.  }
}
\label{Multiple-teacher}
\scalebox{0.87}{
\begin{tabular}{ccccccccc}
\toprule
\textbf{Dataset}                      & \textbf{Teacher} & \textbf{} & \textbf{R@1}  & \textbf{R@5}  & \textbf{R10}  & \textbf{R@100} & \textbf{SumR}  \\ \hline
\multirow{3}{*}{\textbf{TVR}}         & CLIP             &           & 15.3          & 36.0          & 47.5          & 86.0           & 184.8          \\
                                      & BLIP2            &           & 15.1          & 35.6          & 47.4          & 85.5           & 183.6          \\
                                      & CLIP+BLIP2       &           & \textbf{15.7} & \textbf{36.3} & \textbf{47.9} & \textbf{86.6}  & \textbf{186.5} \\ \hline
\multirow{3}{*}{\textbf{ActivityNet}} & CLIP             &           & 8.3           & 25.5          & 38.3          & 77.8           & 149.9          \\
                                      & BLIP2            &           & 8.1           & 25.2          & 38.3          & 77.5           & 149.1          \\
                                      & CLIP+BLIP2       &           & \textbf{8.5}  & \textbf{25.9} & \textbf{38.7} & \textbf{78.8}  & \textbf{151.9} \\ \hline
\multirow{3}{*}{\textbf{Charades}}    & CLIP             &           & 1.9           & 7.1           & 12.3          & 49.8           & 71.1           \\
                                      & BLIP2            &           & 1.9           & 6.8           & 11.3          & 48.6           & 68.7           \\
                                      & CLIP+BLIP2       &           & \textbf{2.1}  & \textbf{7.6}  & \textbf{12.5} & \textbf{50.0}  & \textbf{72.2}          \\
 \bottomrule
\end{tabular}
}
\end{table}

\subsubsection{Extension to Multi-Teacher Distillation}
\label{ssec:multi-teacher}
While our proposed framework has thus far employed single-teacher distillation, it is natural to explore whether the framework can be extended to multi-teacher distillation.
Therefore, we adopt another vision-language pre-training model BLIP-2 \cite{li2023blip} as an extra teacher model. 
As shown in Table \ref{Multiple-teacher}, utilizing BLIP-2 as the teacher model gives comparable performance to the CLIP counterpart. 
Furthermore, with the joint use of CLIP and BLIP-2 as teacher models (their output distributions are fused by simple summation), it brings a further performance boost over the single-teacher distillation.
This extension demonstrates the potential benefits of leveraging multiple vision-language models during training, particularly in scenarios where diverse pre-trained models are readily available.

\subsection{Investigation on Soft Target Alignment}
\subsubsection{Effectiveness of Soft Target Alignment}
To evaluate the effectiveness of our soft targets alignment strategy, we compare the performance of our model against its counterpart that relies on traditional hard targets. As shown in Table~\ref{tab:effect-of-soft-label-module}, the model utilizing soft targets consistently outperforms the hard-target variant across all three datasets, highlighting the advantage of incorporating soft supervision.
These results confirm that effectively mining and leveraging the latent soft relationships between text and video helps the model better capture partial relevance, which is essential for improving performance on the PRVR task.

\begin{table}[tb!]
\renewcommand{\arraystretch}{1.6}
\caption{The effectiveness of soft targets alignment module.
The proposed method with \textit{soft} targets achieves better performance than the counterpart with \textit{hard} targets.
}

\label{tab:effect-of-soft-label-module}
\scalebox{0.87}{
\begin{tabular}{ccccccccc}
\toprule
\textbf{Dataset}                      &  & \textbf{Targets} & \textbf{} & \textbf{R@1}  & \textbf{R@5}  & \textbf{R@10} & \textbf{R@100} & \textbf{SumR}  \\ \hline
\multirow{2}{*}{\textbf{TVR}}         &  & Hard             &           & 14.3         & 34.9          & 45.8          & 84.9           & 179.9          \\
                                      &  & Soft             &           & \textbf{15.3} & \textbf{36.0} & \textbf{47.5} & \textbf{86.0}  & \textbf{184.8} \\ \hline
\multirow{2}{*}{\textbf{ActivityNet}} &  & Hard             &           & 8.0           & 25.0          & 37.5          & 77.1           & 147.6          \\
                                      &  & Soft             &           & \textbf{8.3}  & \textbf{25.5} & \textbf{38.3} & \textbf{77.8}  & \textbf{149.9} \\ \hline
\multirow{2}{*}{\textbf{Charades}}    &  & Hard             &           & 1.5           & 7.0           & 11.1          & 46.7           & 66.4           \\
                                      &  & Soft             &           & \textbf{1.9}  & \textbf{7.1}  & \textbf{12.3} & \textbf{49.8}  & \textbf{71.1}  \\ 
 \bottomrule
\end{tabular}
}
\end{table}

\subsubsection{Qualitative Analysis}
\begin{figure}[tbp!]
        \includegraphics[width=1\linewidth]{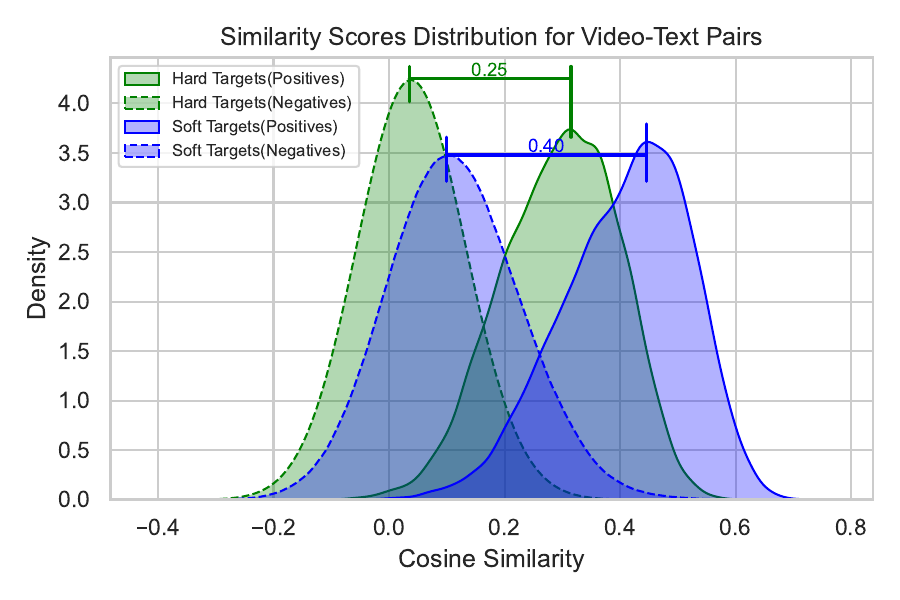}
	\vspace{-8mm}
        \caption{Similarity distribution of positive and negative video-text pairs. Using soft targets enhances the separation between positive and negative samples, leading to more discriminative representations. 
	}
	\label{fig:Similarity Scores}

\end{figure}

Fig. \ref{fig:Similarity Scores} illustrates the similarity distribution of video-text pairs on the validation split of the ActivityNet dataset, comparing our model with soft targets and the counterpart using hard targets.
The distributions clearly show that our method produces a larger separation between the positive and negative sample distributions, with the center distance increasing from 0.25 to 0.40.
This increased margin indicates that the soft targets alignment module enhances the model's ability to distinguish between positive and negative pairs, leading to more discriminative representations. The reduced overlap between the positive and negative distributions further supports the effectiveness of our approach in modeling fine-grained video-text relevance, which is crucial for improving retrieval performance in the PRVR task.

\subsubsection{Influence of decay strategies in soft targets construction}
In this experiment, we further investigate the effect of different decay strategies for the parameters $\alpha$ and $\beta$, which control the dynamic construction of soft targets. Similar to the earlier analysis of $w$, we compare exponential, sigmoid, and linear decay schedules, along with a fixed-parameter baseline. The results are shown in Fig.~\ref{fig:soft paratmers dynamic}.
Across all datasets, the three decay-based strategies consistently outperform the fixed baseline, highlighting the benefit of dynamically adjusting soft targets during training. Additionally, while the overall trends are similar across strategies, the sigmoid decay generally achieves slightly better or more stable results, particularly by balancing early hard-target supervision with a gradual emphasis on softer targets as training progresses.
In particular, setting $\alpha$ to 0 disables the sample-wise dynamic target selection, causing the model to rely entirely on soft targets without any hard supervision. Similarly, setting $\beta$ to 0 removes the influence of human-annotated labels in the score-wise dynamic target refinement, relying solely on predicted relevance scores. In both cases, the model exhibits consistently degraded performance.
These results demonstrate the importance of maintaining a balance between soft and hard targets, as well as between predicted and annotated signals. Taken together, these findings validate the necessity of a well-designed decay strategy.
Accordingly, we adopt the sigmoid decay function as the default configuration for both $\alpha$ and $\beta$ in our framework.

\begin{figure*}[tb!]
\centering
\vspace{-1mm}
\renewcommand\thesubfigure{}
\captionsetup[subfloat]{labelformat=empty}

\subfloat[]{\includegraphics[width=0.31\linewidth]{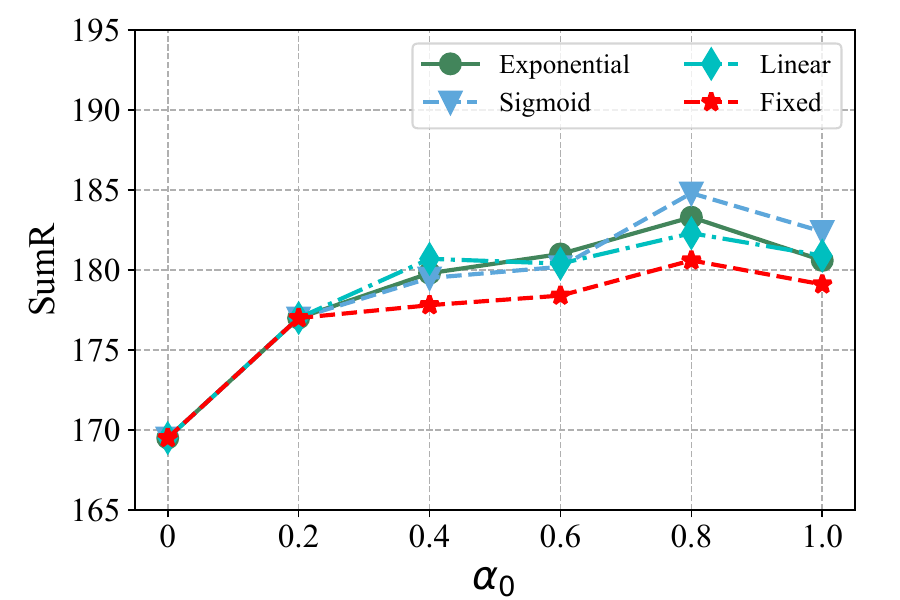}}\hfill
\subfloat[]{\includegraphics[width=0.31\linewidth]{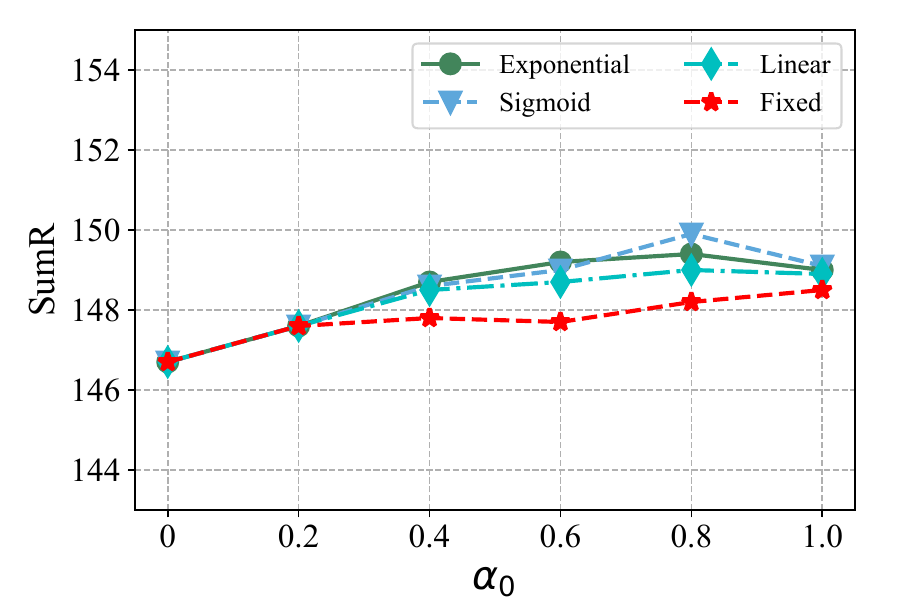}}\hfill
\subfloat[]{\includegraphics[width=0.31\linewidth]{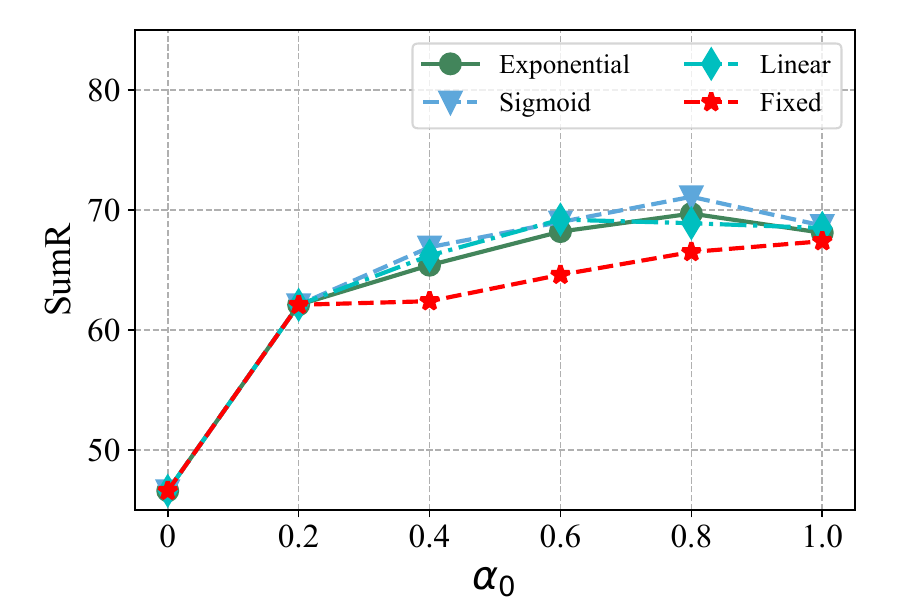}}

\vspace{-4mm}

\renewcommand\thesubfigure{\alph{subfigure}}
\captionsetup[subfloat]{labelformat=parens, labelsep=space}
\setcounter{subfigure}{0}

\subfloat[TVR]{\includegraphics[width=0.31\linewidth]{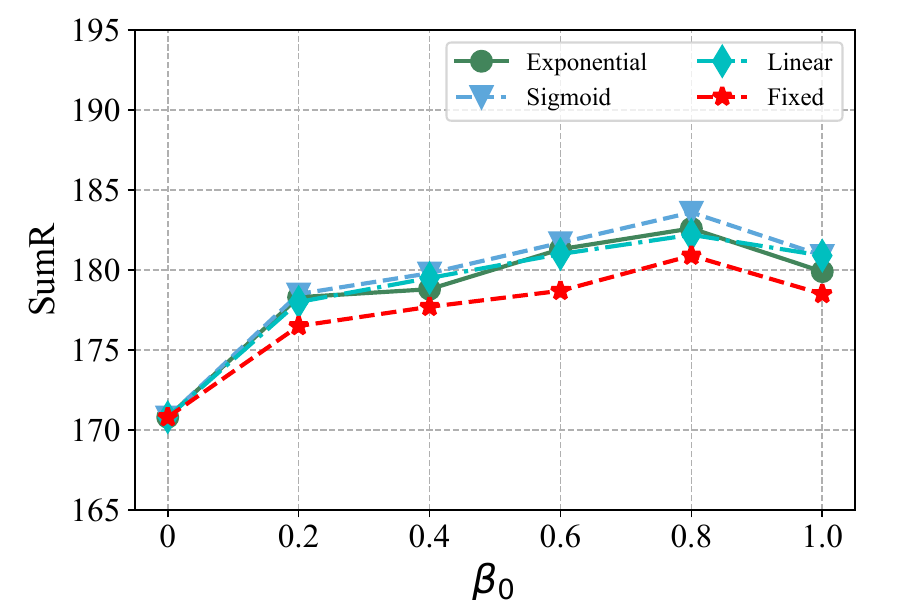}}\hfill
\subfloat[ActivityNet Captions]{\includegraphics[width=0.31\linewidth]{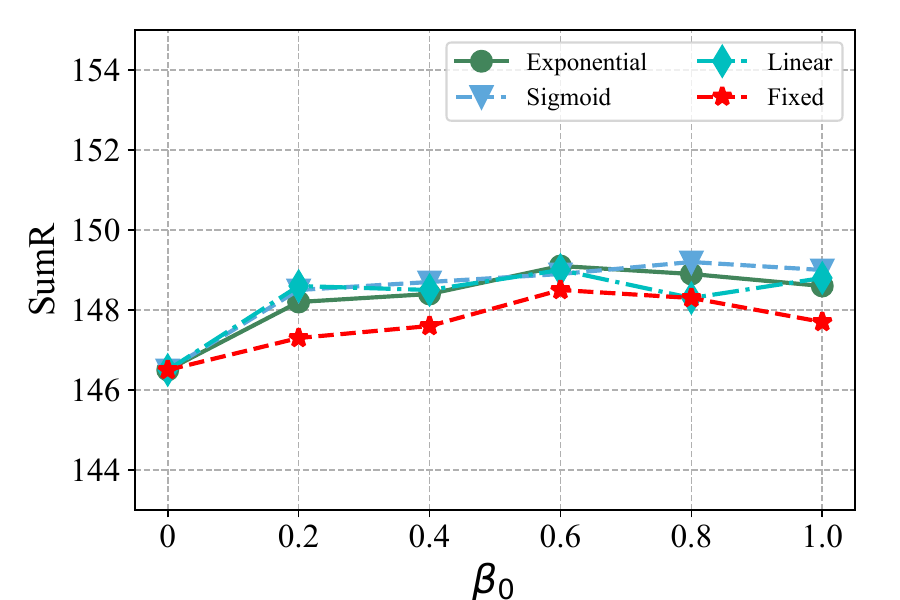}}\hfill
\subfloat[Charades-STA]{\includegraphics[width=0.31\linewidth]{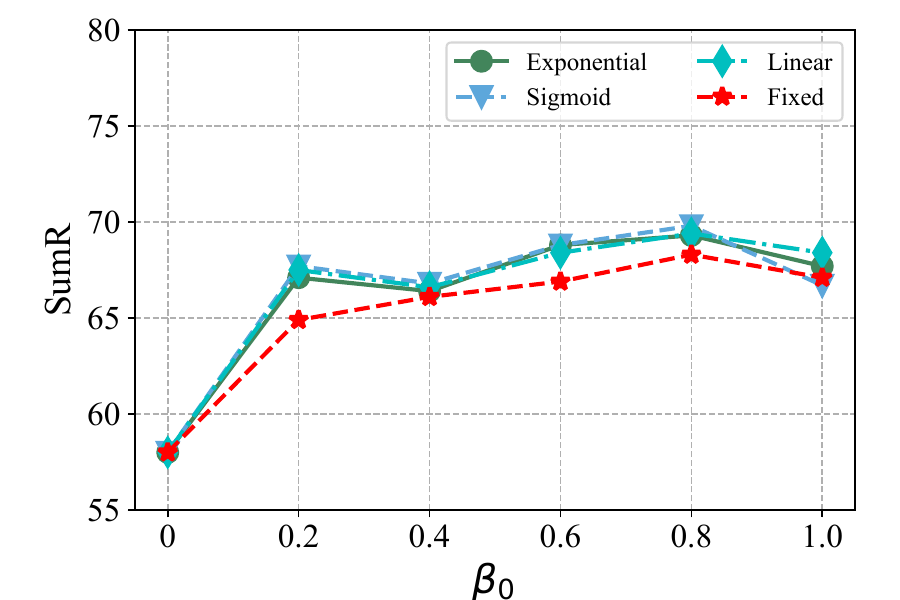}}

\vspace{-1mm}
\caption{The influence of decay strategies in our dynamic soft targets construction.
The three decay strategies demonstrate comparable performance.}
\vspace{-3mm}
\label{fig:soft_paratmers_dynamic}
\end{figure*}

\begin{figure*}[tb!]
        \includegraphics[width=1\linewidth]{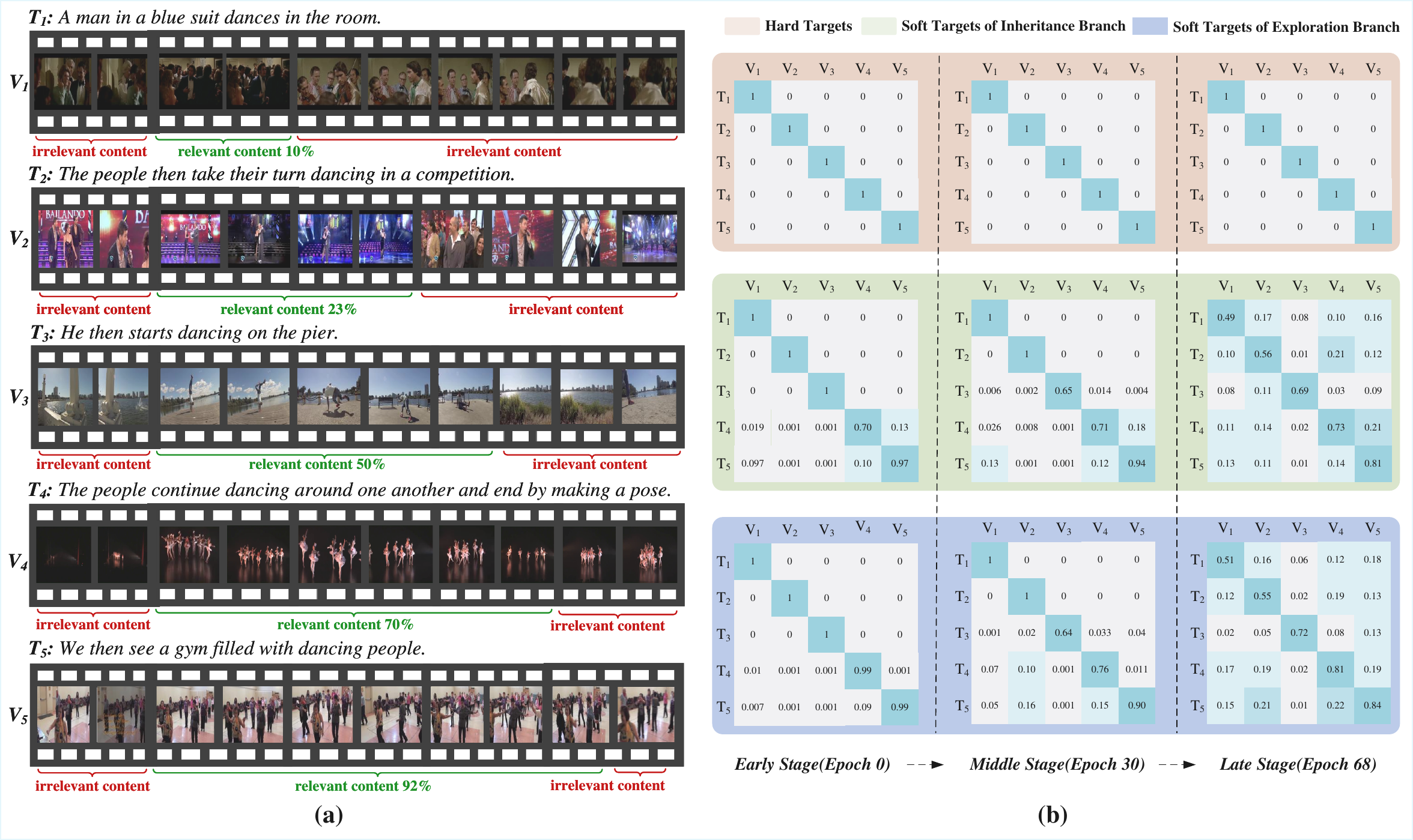}
	\vspace{-4mm}
        \caption{The visualization results on the ActivityNet dataset.
        We present the soft targets output by our method for 5 video-text samples with different M/V ratios, and the dynamic evolution of soft targets during the training stage. }
	\label{fig:soft examples}
\vspace{-3mm}
\end{figure*}

\subsection{Visualization of Soft Targets}

To further illustrate the effectiveness of our dynamic soft target construction, we visualize representative examples from the ActivityNet dataset in Fig.\ref{fig:soft examples}. Specifically, we select five semantically similar video-text pairs with varying moment-to-video (M/V) ratios to analyze how our model captures partial relevance throughout training. As shown in Fig.\ref{fig:soft examples}(a), each video contains a different proportion of query-relevant content. Notably, certain video-text pairs exhibit partial but non-trivial semantic overlap despite being annotated as negative under traditional hard target schemes.

In Fig.~\ref{fig:soft examples}(b), we present the evolution of soft target similarity matrices across three training stages (early, middle, late epochs), as predicted by both the inheritance and exploration branches. Several key observations emerge: (1) At the early stage (Epoch 0), both branches largely mirror hard targets, resulting in sharply binarized similarity scores. (2) As training progresses to the middle stage (Epoch 30), the model begins to adjust similarity scores based on the nuanced degrees of relevance between video-text pairs. For example, the soft target scores between $T_5$ and $V_4$ (which are partially relevant) gradually increase to reflect their semantic alignment (reaching scores of 0.14 and 0.22 in the exploration branch). Similar trends are observed for other partially relevant negative pairs, such as $T_2$ with $V_4$, and $T_1$ with $V_5$. (3) By the late stage (Epoch 68), the soft targets for positive pairs progressively stabilize at levels consistent with their M/V ratios, while the targets for partially relevant negatives continue to reflect meaningful partial similarity rather than being constrained to zero.
These results validate that our dynamic soft target construction module enables the model to transcend rigid binary supervision and effectively capture fine-grained semantic relationships, particularly for complex scenarios involving partial relevance.

\section{Conclusions}
In this paper, we have investigated the meaningful but challenging text-to-video subtask of PRVR from a new perspective of knowledge distillation. 
Besides, we have argued that conventional binary hard targets, which treat all positive samples equally, are suboptimal for the PRVR task, as they fail to reflect the varying degrees of partial relevance in untrimmed videos.
To overcome these limitations, a novel framework, \textit{i.e.}, DL-DKD++, has been proposed to distill the generalization knowledge from the large-scale vision-language pre-trained model to a task-specific network.
The framework further benefits from joint learning with dynamic soft targets, allowing for more reasonable supervision.
Extensive experiments on three datasets support the following conclusions:
(1) Dual learning of an inheritance branch and an exploration branch is essential for learning generalizable PRVR knowledge.
(2) Dynamic knowledge distillation further enhances performance, especially when the teacher model performs moderately.
(3) Incorporating dynamic soft targets further enhances the model’s ability to learn fine-grained partial relevance.

\vspace{3mm}
\textbf{Acknowledgements}.
This work was supported by the the National Natural Science Foundation of China (No. 62472385), ``Pioneer" and ``Leading Goose" R\&D Program of Zhejiang (No. 2024C01110), Young Elite Scientists Sponsorship Program by CAST (No. 2022QNRC001), Zhejiang Provincial Natural Science Foundation (No. LZ23F020004), Fundamental Research Funds for the Provincial Universities of Zhejiang (No. FR2402ZD).

{\small
\bibliographystyle{ieee_fullname}
\bibliography{egbib}
}

\vspace{-10mm}
\begin{IEEEbiography}[{\includegraphics[width=1in,height=1.25in,clip,keepaspectratio]{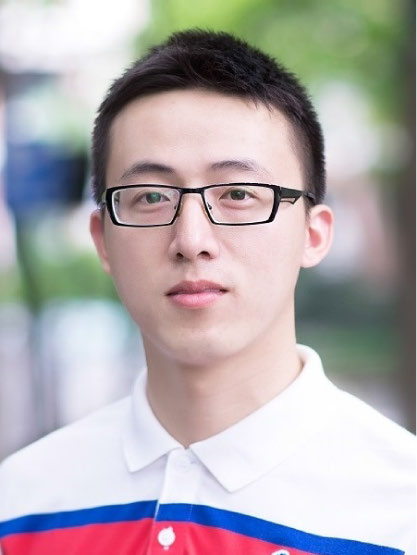}}]{Jianfeng Dong}
received the B.E. degree in software engineering from the Zhejiang University of Technology, China, in 2013, and the Ph.D. degree in computer science from Zhejiang University, China, in 2018. He is currently a Research Professor with the College of Computer Science and Technology, Zhejiang Gongshang University, Hangzhou, China. His research interests include multimedia understanding, retrieval, and recommendation. He was awarded the ACM Multimedia Grand Challenge Award and was selected into the Young Elite Scientists Sponsorship Program by the China Association for Science and Technology.
\end{IEEEbiography}

\vspace{-10mm}
\begin{IEEEbiography}
[{\includegraphics[width=1in,height=1.25in,clip,keepaspectratio]{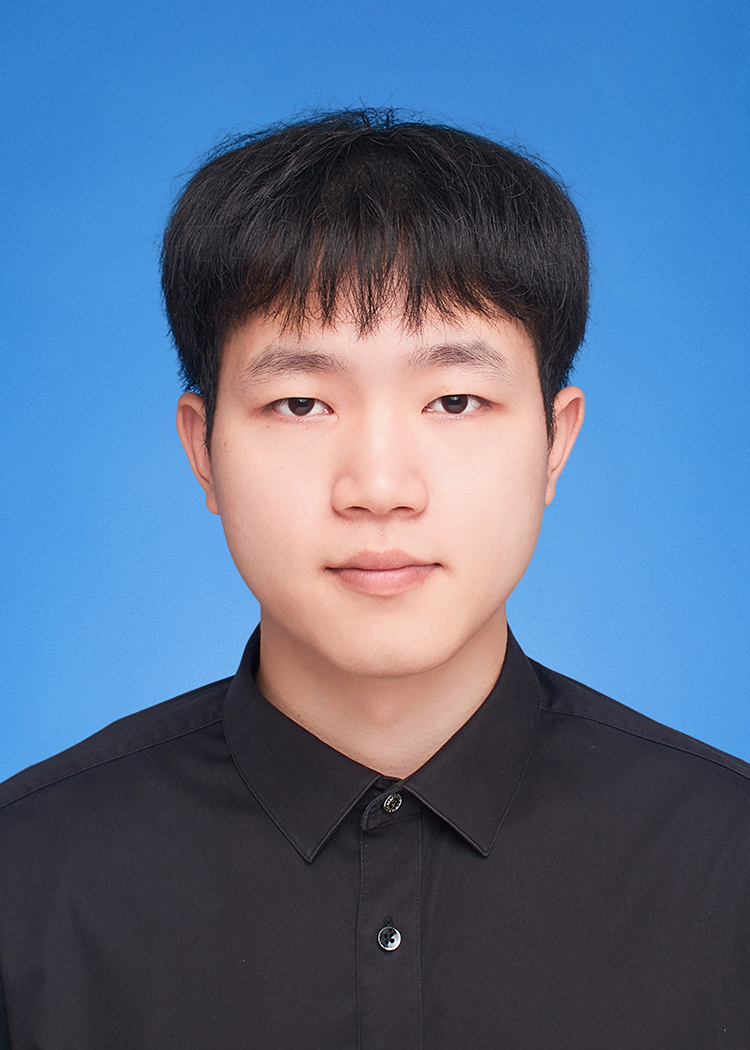}}]{Lei Huang}
received his bachelor's degree in computer science and technology from Chongqing Jiaotong University, China, in 2023. Currently, he is pursuing a master's degree in computer technology from Zhejiang Gongshang University. His research interests include multimodal learning.
\end{IEEEbiography}

\vspace{-10mm}
\begin{IEEEbiography}
[{\includegraphics[width=1in,height=1.25in,clip,keepaspectratio]{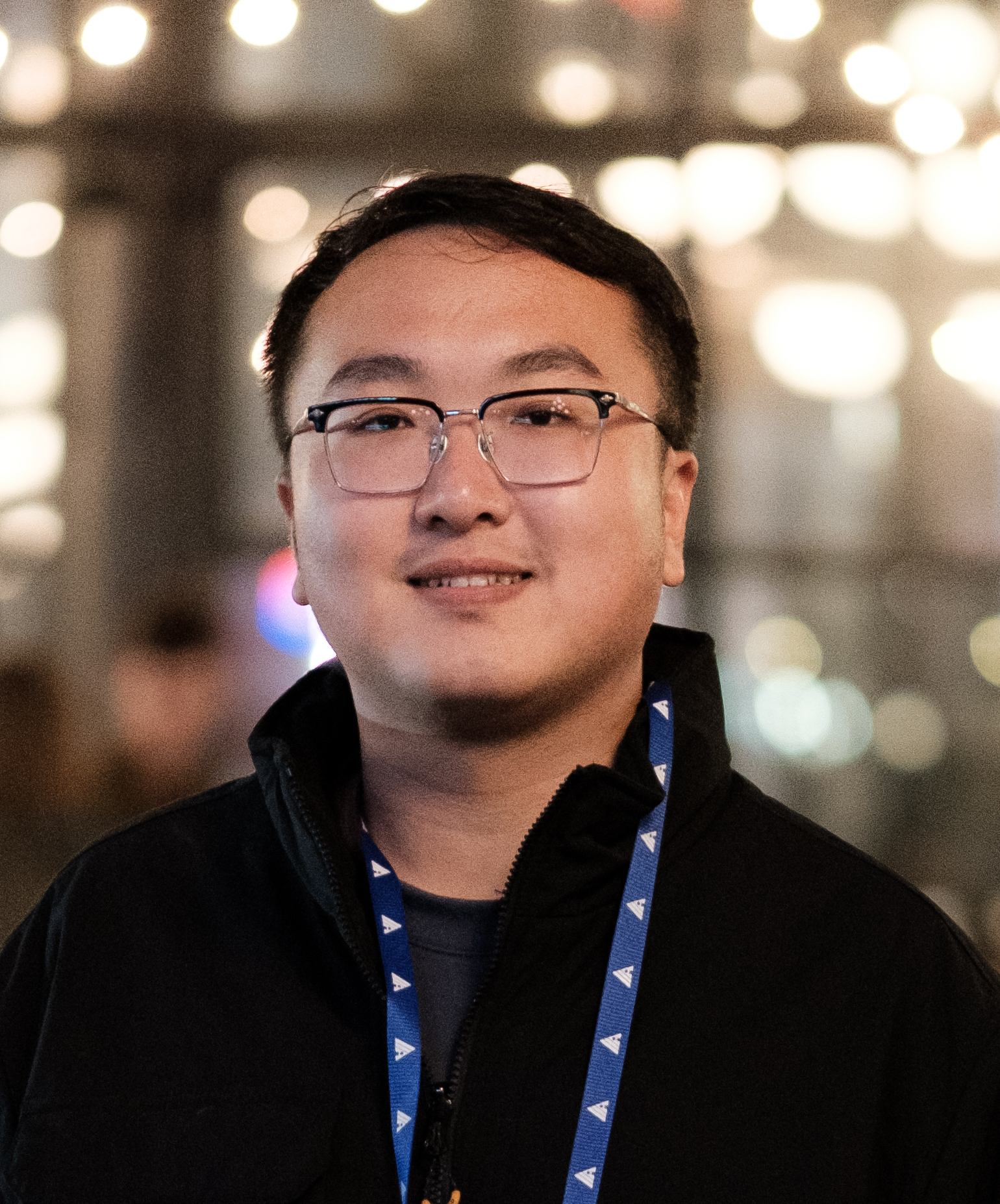}}]{Daizong Liu} received the M.S. degree in Electronic Information and Communication of Huazhong University of Science and Technology in 2021. He is currently working toward the Ph.D. degree at Wangxuan Institute of Computer Technology of Peking University. His research interests include 3D adversarial attacks, multi-modal learning, LVLM robustness, etc. He has published more than 40 papers in refereed conference proceedings and journals such as TPAMI, NeurIPS, CVPR, ICCV, ECCV, SIGIR, AAAI. He regularly serves on the program committees of top-tier AI conferences such as NeurIPS, ICML, ICLR, CVPR, ICCV and ACL.
\end{IEEEbiography}

\vspace{-10mm}
\begin{IEEEbiography}
[{\includegraphics[width=1in,height=1.25in,clip,keepaspectratio]{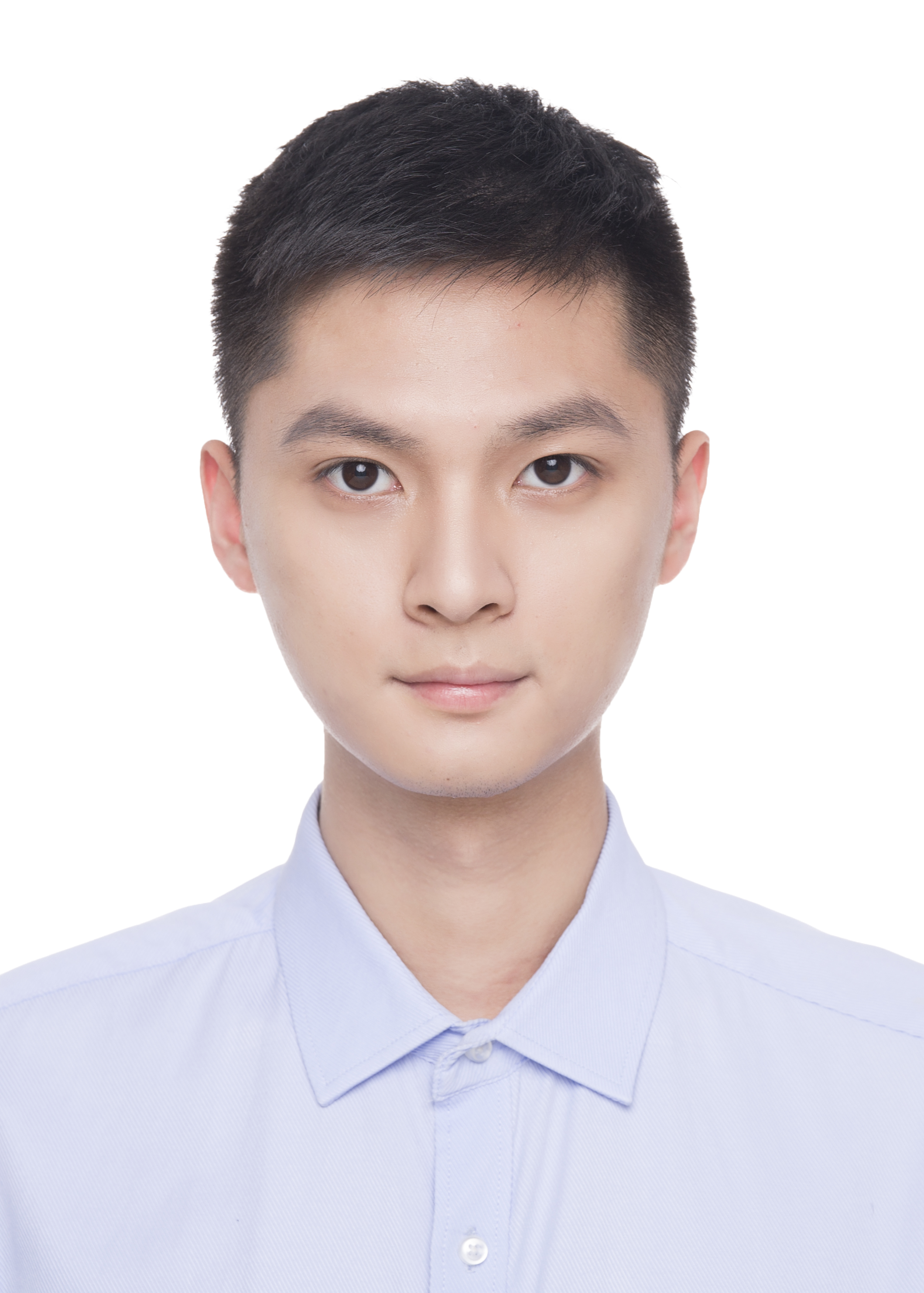}}]{Xianke Chen}
received the B.E. degree in network engineering from Zhejiang Gongshang University, Hangzhou,
China, in 2020, and the M.E. degree from the College of Computer Science and Technology, Zhejiang
Gongshang University, Hangzhou, China, in 2023.
He is currently pursuing the Ph.D. degree with
the Big data statistics, Zhejiang Gongshang University. His research interests include multi-modal learning.
\end{IEEEbiography}

\vspace{-10mm}
\begin{IEEEbiography}[{\includegraphics[width=1in,height=1.25in,clip,keepaspectratio]{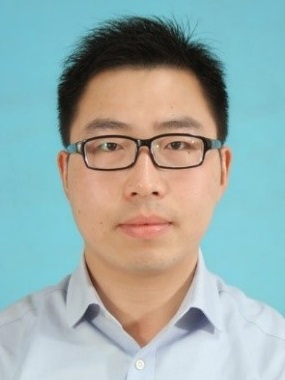}}]{Xun Yang} received the Ph.D. degree from the School of Computer and Information, Hefei University of Technology, China, in 2018. He is currently a postdoctoral research fellow with the NExT++ Research Center, National University of Singapore, Singapore. His current research interests include information retrieval, multimedia content analysis, and computer vision. He has served as the PC member and the invited reviewer for top-tier conferences and prestigious journals including ACM MM, IJCAI, AAAI, the ACM Transactions on Multimedia Computing, Communications, and Applications, IEEE Transactions on Neural Networks and Learning Systems, IEEE Transactions on Knowledge and Data Engineering, and IEEE Transactions on Circuits and Systems for Video Technology.
\end{IEEEbiography}

\vspace{-10mm}
\begin{IEEEbiography}[{\includegraphics[width=1in,height=1.25in,clip,keepaspectratio]{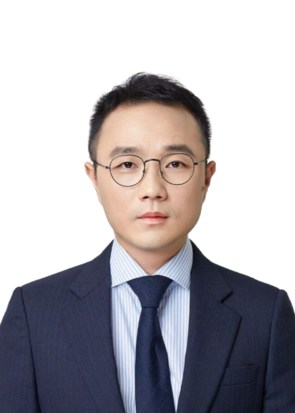}}]{Changting Lin}
received his Ph.D. in Computer Science from Zhejiang University in 2018. He is currently a researcher at the Binjiang Institute of Zhejiang University, China, and also a co-founder of Gentel.io. His research interests cover the fields of artificial intelligence and artificial intelligence security.
\end{IEEEbiography}

\vspace{-10mm}
\begin{IEEEbiography}[{\includegraphics[width=1in,height=1.25in,clip,keepaspectratio]{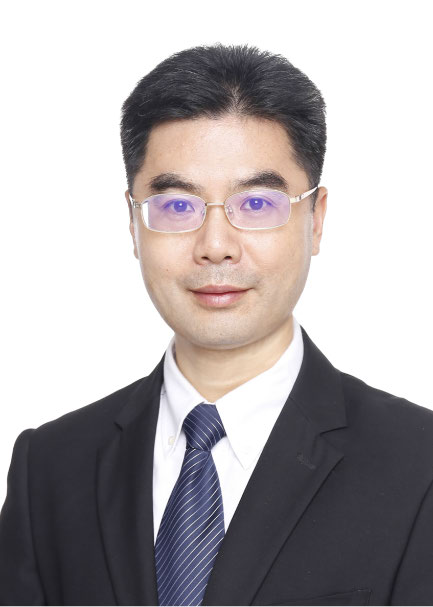}}]{Xun Wang} (Member, IEEE) received the B.S. degree in mechanics and the Ph.D. degrees in computer science from Zhejiang University, Hangzhou, China, in 1990 and 2006, respectively. He is currently a professor at the School of Computer Science and Information Engineering, Zhejiang Gongshang University, China. His current research interests include mobile graphics computing, image/video processing, pattern recognition, intelligent information processing and visualization. He is also a member of the ACM, and a senior member of the CCF.
\end{IEEEbiography}

\vspace{-10mm}
\begin{IEEEbiography}[{\includegraphics[width=1in,height=1.25in,clip,keepaspectratio]{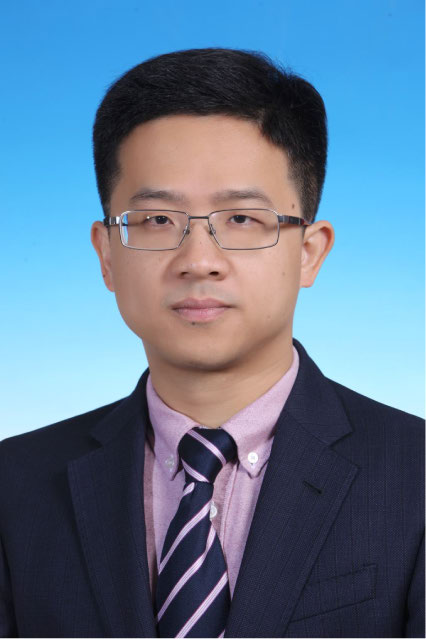}}]{Meng Wang} (Fellow, IEEE) received the B.E. and Ph.D. degrees in the special class for the Gifted Young and the Department of Electronic Engineering and Information Science, University of Science and Technology of China (USTC), Hefei, China, in 2003 and 2008, respectively. He is currently a professor with the Hefei University of Technology, China. His current research interests include multimedia content analysis, computer vision, and pattern recognition. He has authored more than 200 book chapters, journal and conference papers in these areas. He is the recipient of the ACM SIGMM Rising Star Award 2014. He is an associate editor of the IEEE Transactions on Knowledge and Data Engineering (IEEE TKDE), IEEE Transactions on Circuits and Systems for Video Technology (IEEE TCSVT), IEEE Transactions on Multimedia (IEEE TMM),and IEEE Transactions on Neural Networks and Learning Systems (IEEE TNNLS).
\end{IEEEbiography}

\end{document}